\documentclass{article}

\PassOptionsToPackage{numbers, compress}{natbib}
\usepackage[final]{neurips_2025}

\usepackage[utf8]{inputenc} % allow utf-8 input
\usepackage[T1]{fontenc}    % use 8-bit T1 fonts
\usepackage{hyperref}       % hyperlinks
\usepackage{url}            % simple URL typesetting
\usepackage{booktabs}       % professional-quality tables
\usepackage{amsfonts}       % blackboard math symbols
\usepackage{nicefrac}       % compact symbols for 1/2, etc.
\usepackage{microtype}      % microtypography
\usepackage{xcolor}         % colors

\usepackage{bm}
\usepackage{amsmath}
\usepackage{amssymb}
\usepackage{colortbl}
\usepackage{graphicx}
\usepackage{enumitem}
\usepackage{multirow}
\usepackage{wrapfig}     % wrap text around figure
\newcommand{\ie}{\textit{i}.\textit{e}.}
\newcommand{\eg}{\textit{e}.\textit{g}.}
\newcommand{\vs}{\textit{vs}.}

\newcommand{\best}{\cellcolor{orange}}
\newcommand{\sbest}{\cellcolor{pink}}
\newcommand{\tbest}{\cellcolor{yellow}}
\title{VA-GS: Enhancing the Geometric Representation of Gaussian Splatting via View Alignment}

\author{
Qing Li$^1$
\quad
Huifang Feng$^2$\thanks{Corresponding author}
\quad
Xun Gong$^1$
\quad
Yu-Shen Liu$^3$\\
$^1$ Southwest Jiaotong University, Chengdu, China\\
$^2$ Xihua University, Chengdu, China
\quad
$^3$ Tsinghua University, Beijing, China\\
{\tt\small qingli@swjtu.edu.cn} \quad {\tt\small fhf@xhu.edu.cn} \quad {\tt\small xgong@swjtu.edu.cn} \quad {\tt\small liuyushen@tsinghua.edu.cn}
}

\begin{document}

\maketitle

\begin{abstract}
    3D Gaussian Splatting has recently emerged as an efficient solution for high-quality and real-time novel view synthesis.
    However, its capability for accurate surface reconstruction remains underexplored.
    Due to the discrete and unstructured nature of Gaussians, supervision based solely on image rendering loss often leads to inaccurate geometry and inconsistent multi-view alignment.
    In this work, we propose a novel method that enhances the geometric representation of 3D Gaussians through view alignment (VA).
    Specifically, we incorporate edge-aware image cues into the rendering loss to improve surface boundary delineation.
    To enforce geometric consistency across views, we introduce a visibility-aware photometric alignment loss that models occlusions and encourages accurate spatial relationships among Gaussians.
    To further mitigate ambiguities caused by lighting variations, we incorporate normal-based constraints to refine the spatial orientation of Gaussians and improve local surface estimation.
    Additionally, we leverage deep image feature embeddings to enforce cross-view consistency, enhancing the robustness of the learned geometry under varying viewpoints and illumination.
    Extensive experiments on standard benchmarks demonstrate that our method achieves state-of-the-art performance in both surface reconstruction and novel view synthesis.
    The source code is available at \textcolor{red}{\href{https://github.com/LeoQLi/VA-GS}{https://github.com/LeoQLi/VA-GS}}.
    % We will release our code to support future research.
\end{abstract}

%%%%%%%%%%%%%%%%%%%%%%%%%%%%%%%%%%%%%%%%%%%%%%%%%%%%%%%%%%%%
\section{Introduction}

Accurate surface reconstruction from multi-view images is a long-standing problem in computer vision, fundamental to applications such as 3D modeling, AR/VR, and robotics.
Recently, 3D Gaussian Splatting (3DGS) has emerged as a powerful explicit representation for real-time novel view synthesis, demonstrating impressive rendering quality and speed by modeling scenes as collections of semi-transparent 3D Gaussian primitives. %and rendering them via splatting-based techniques.
However, despite its rendering advantages, 3DGS remains limited in its ability to recover accurate and detailed geometry, especially when supervision is derived solely from RGB images.
This limitation stems from the inherent discrete and unstructured nature of Gaussians, which makes it difficult to enforce global surface consistency or capture fine geometric details, particularly under complex illumination and along object boundaries.

\begin{wrapfigure}{r}{0.6\textwidth}
    \vspace{-.5cm}
    \centering
    \includegraphics[width=\linewidth]{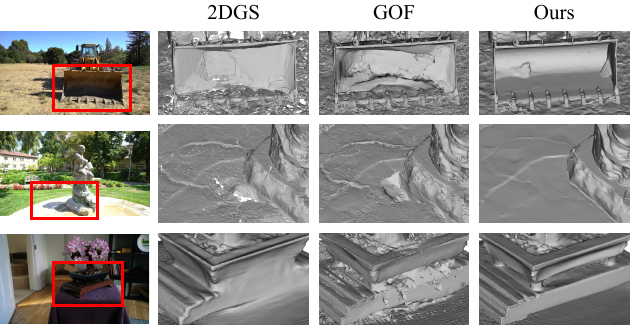}  \vspace{-0.6cm}
    \caption{
        Our method addresses illumination and boundary artifacts that previous methods fail to resolve.
    }
    \label{fig:intro}
    \vspace{-0.1cm}
\end{wrapfigure}

Existing methods have attempted to enhance the geometric capabilities of Gaussian splatting.
For example, SuGaR~\cite{guedon2024sugar} constructs a density field from Gaussians and extracts meshes via level-set searching, but it struggles with large smooth surfaces and is computationally expensive.
2DGS~\cite{huang20242d} models scenes using 2D oriented planar Gaussian disks, which inherently represent surfaces and provide view-consistent geometry.
However, 2DGS has difficulty reconstructing background geometry and often produces incomplete or distorted surfaces in complex or unbounded scenes.
GOF~\cite{yu2024gaussian} constructs an opacity field from Gaussians and extracts surfaces using Marching Tetrahedra~\cite{doi1991efficient}, yielding adaptive mesh resolution without volumetric fusion.
Nonetheless, thin structures can be lost and strong lighting contrasts still cause artifacts.
GS-Pull~\cite{zhang2024neural} integrates a neural signed distance field (SDF), dynamically pulling Gaussians toward the zero-level set of the learned SDF.
While this improves surface completeness, it introduces additional network complexity, produces overly smooth surfaces, and primarily focuses on foreground object reconstruction.
PGSR~\cite{chen2024pgsr} fits Gaussians to local planar hypotheses and uses unbiased depth rendering to improve geometric accuracy.
However, it does not fully resolve the challenges posed by complex lighting and remains sensitive to boundary ambiguities in non-planar regions.
Overall, previous methods have introduced geometric regularizers or hybrid representations and achieved significant progress.
However, they still struggle to address two persistent challenges: illumination-induced artifacts (\eg, shadows and specular highlights) and accurate surface boundary delineation, as shown in Fig.~\ref{fig:intro}.
Illumination effects distort photometric losses, while ambiguous boundaries often result in geometry drift or holes.

In this work, we propose a novel method for accurate and detailed surface reconstruction by enhancing the geometric representation of 3D Gaussians.
We address the limitations of previous methods by incorporating multi-faceted geometric constraints and structural priors.
Our approach introduces geometry-aware constraints guided by image edges, multi-view alignment that considers visibility and occlusion, and robust priors derived from surface normals and deep image features to mitigate the effects of lighting variations and boundary ambiguities.
Specifically, we enhance the standard rendering loss with edge-aware image cues, which sharpen surface boundaries in the 2D projection space of Gaussian splats, resulting in clearer and more precise delineations in rendered images.
To enforce geometric consistency across views, we introduce a multi-view photometric alignment loss that explicitly accounts for visibility and occlusions, encouraging accurate spatial relationships among 3D Gaussians and improving boundary localization.
To further reduce ambiguity caused by lighting, we introduce normal-based alignment to constrain the spatial orientation of Gaussians, ensuring reliable surface estimation.
Additionally, we leverage high-dimensional image features to enforce cross-view consistency, improving robustness to viewpoint and lighting variations.
These innovations significantly reduce the impact of complex illumination and boundary ambiguity, enabling accurate surface reconstruction in challenging scenes.
Experiments on standard benchmarks demonstrate that our method achieves state-of-the-art performance in both surface reconstruction and novel view synthesis.
Our contributions are summarized as follows.
\begin{itemize}[leftmargin=*]
\setlength{\itemsep}{1pt}
\setlength{\parsep}{0pt}
\setlength{\parskip}{0pt}
\vspace{-0.2cm}
\item Incorporating edge information and visibility-aware multi-view alignment to enhance surface boundary delineation and improve geometric consistency.
\item Aligning the robust priors based on normals and deep image features to mitigate illumination-induced artifacts and increase reconstruction accuracy.
\item State-of-the-art results on standard benchmarks, demonstrating the effectiveness of our method in both surface reconstruction and novel view synthesis.
\end{itemize}

%%%%%%%%%%%%%%%%%%%%%%%%%%%%%%%%%%%%%%%%%%%%%%%%%%%%%%%%%%%%
\section{Related Work}

\textbf{View Synthesis and Gaussian Splatting.}
Neural Radiance Fields (NeRF)~\cite{mildenhall2020nerf} pioneered high-fidelity novel view synthesis by representing a scene as a continuous volumetric density and view-dependent color field, optimized via differentiable volume rendering.
Subsequent works accelerated training and rendering through hybrid representations such as multi-resolution hash grids~\cite{muller2022instant}, explicit voxel or sparse tensor grids~\cite{sun2022direct,barron2022mip}, and learned feature planes~\cite{yu2021plenoctrees}.
However, these volumetric methods still entail high memory and computational costs.
3DGS~\cite{kerbl20233d} departs from dense volumes by modeling a scene as a sparse cloud of anisotropic 3D Gaussians.
Follow-up work has enhanced visual fidelity through anti-aliasing and level-of-detail control~\cite{yu2024mip,song2024sa}, improved training speed and robustness under sparse views using density regularization and learned radiance priors~\cite{yang2024spectrally,niemeyer2024radsplat}, and extended 3DGS to dynamic scenes~\cite{luiten2023dynamic,yang2023gs4d}, relighting~\cite{gao2023relightable}, and animation~\cite{ye2023animatable}.
Geometry-aware variants such as FatesGS~\cite{huang2025fatesgs}, DNGaussian~\cite{li2024dngaussian} and GeoGaussian~\cite{li2024geogaussian} address sparse-view and textureless regions, while methods like Instantsplat~\cite{fan2024instantsplat} and Scaffold-GS~\cite{lu2024scaffold} accelerate convergence by leveraging pretraining or hybrid implicit-explicit designs.
Despite these advances, most 3DGS variants primarily emphasize appearance quality and lack mechanisms to enforce explicit surface geometry, motivating dedicated reconstruction techniques.

\textbf{Surface Reconstruction with Gaussians.}
Extracting accurate surfaces from a 3DGS representation is challenging due to its unstructured nature and supervision based solely on RGB signals.
Early approaches convert Gaussians into volumetric density or opacity fields:
SuGaR~\cite{guedon2024sugar} builds a density field and applies level-set search with Poisson reconstruction~\cite{kazhdan2006poisson}, but it struggles to recover large, smooth surfaces;
GOF~\cite{yu2024gaussian} accumulates per-view alpha values into an opacity volume and extracts iso-surfaces with Marching Tetrahedra~\cite{doi1991efficient}, achieving adaptive resolution but often missing fine, thin structures under high lighting contrast.
Other methods project Gaussians into oriented 2D disks (surfels) and fuse via Truncated Signed Distance Function (TSDF) fusion~\cite{newcombe2011kinectfusion} or Poisson reconstruction.
2DGS~\cite{huang20242d} and GSurfels~\cite{dai2024high} improve local alignment but tend to introduce distortions in unbounded scenes and result in incomplete background geometry.
PGSR~\cite{chen2024pgsr} fits Gaussians to planar patches and adds multi-view photometric and geometric regularization, excelling on planar man-made scenes but remaining sensitive to non-planar boundaries.
More recent works~\cite{yu2024gsdf,zhang2024neural,lyu20243dgsr,baixin2024gsurf,Li2024MonoGSDFEM,li2025gaussianudf} integrate Signed Distance Fields (SDF) to guide Gaussian placement.
GSDF~\cite{yu2024gsdf} and 3DGSR~\cite{lyu20243dgsr} jointly optimize a neural SDF branch alongside Gaussian parameters using volume-rendered depth and normal supervision, which improves surface smoothness but requires additional network branches.
GS-Pull~\cite{zhang2024neural} leverages SDF gradients to pull Gaussians toward the zero-level set, enhancing alignment at the cost of limiting object-level reconstruction and producing overly smooth results.
Methods that incorporate depth or normal estimators~\cite{chen2024vcr,zhang2024rade,turkulainen2024dn,wolf2024surface,wang2024gaussurf} impose priors on Gaussians but rely on TSDF fusion's fixed resolution or Poisson reconstruction's sensitivity to noisy inputs, and often struggle under varying illumination or around complex geometric boundaries.
Our approach enforces view consistency through multi-faceted constraints during Gaussian optimization, enabling high-fidelity mesh extraction even under challenging lighting and boundary conditions.
% GausSurf~\cite{wang2024gaussurf} integrates patch-match based multi-view stereo guidance in texture-rich regions and normal priors in textureless areas to iteratively refine Gaussian geometry.

%%%%%%%%%%%%%%%%%%%%%%%%%%%%%%%%%%%%%%%%%%%%%%%%%%%%%%%%%%%%
\section{Preliminaries}

3DGS~\cite{kerbl20233d} explicitly represents a scene as a collection of anisotropic 3D Gaussians, which can be rendered to images from arbitrary viewpoints using a splatting-based rasterization technique~\cite{zwicker2001ewa}.
Specifically, each 3D Gaussian $\mathbb{G}$ is defined as:
\begin{equation}
    \mathbb{G}(x) = \exp\left( -0.5 (x - \bm{\mu})^T \bm{\Sigma}^{-1} (x - \bm{\mu}) \right),
\end{equation}
where $\bm{\mu}$ is the Gaussian center and $\bm{\Sigma}$ is its covariance matrix.
For novel-view rendering, the color at pixel $\bm{p}$ is obtained by compositing $K$ ordered Gaussian splats using point-based $\alpha$-blending, \ie,
\begin{equation} \label{eq:colorblending}
    \bm{C}(\bm{p}) = \sum_{i=1}^K \bm{c}_i \alpha_i \prod_{j=1}^{i-1}\left(1-\alpha_j\right),
\end{equation}
where $\alpha_i$ denotes the pixel translucency determined by the learned opacity of the $i$-th Gaussian kernel and its projected footprint at pixel $\bm{p}$.
The view-dependent color $\bm{c}_i$ is encoded using spherical harmonics associated with each Gaussian.
In addition to color, Eq.~\eqref{eq:colorblending} is similarly used to render per-pixel normals and depths by replacing $\bm{c}_i$ with the corresponding normal or depth value.

\textbf{Normal and Depth Estimation from Gaussians.}
The covariance matrix $\bm{\Sigma} \in \mathbb{R}^{3 \times 3}$ of a 3D Gaussian can be decomposed into a rotation matrix $\bm{R}$ and a scaling matrix $\bm{S}$, \ie,
$\bm{\Sigma} = \bm{R} \bm{S} \bm{S}^\top \bm{R}^\top$,
where $\bm{R}$ contains the three orthogonal eigenvectors, and $\bm{S}$ encodes the scale along these directions.
This decomposition resembles an ellipsoid representation: the eigenvectors define the axes of the ellipsoid, while the scale values correspond to the axis lengths.
As optimization progresses, the initially spherical Gaussian flattens and approaches a plane~\cite{jiang2023gaussianshader}.
We take the direction corresponding to the smallest scale factor as the normal $\bm{n}$ of the Gaussian.
The distance from the local plane to the camera center is then computed as
$\bm{d} = (\bm{R}_c^\top (\bm{\mu} - \bm{T}_c))^\top (\bm{R}_c^\top \bm{n})$,
where $\bm{R}_c$ is the rotation from the camera to the world frame, and $\bm{T}_c$ is the camera center in world coordinates.
Given the normal and distance, the depth is obtained by intersecting the viewing ray with the local plane:
$\bm{z} = \bm{d} / (\bm{R}_c^\top \bm{n} ~\bm{K}^{-1} \bar{\bm{p}})$,
where $\bar{\bm{p}}$ is the homogeneous coordinate of the pixel (we use $\bm{p}$ to denote both the homogeneous and 2D pixel coordinates for simplicity), and $\bm{K}$ is the intrinsic matrix of the camera.
Finally, the per-pixel distance, depth, and normal maps under the current viewpoint are rendered using $\alpha$-blending as defined in Eq.~\eqref{eq:colorblending}, where the attribute color $\bm{c}_i$ is replaced with the corresponding Gaussian attributes.

\begin{figure}[t]
    \centering
    \includegraphics[width=.9\linewidth]{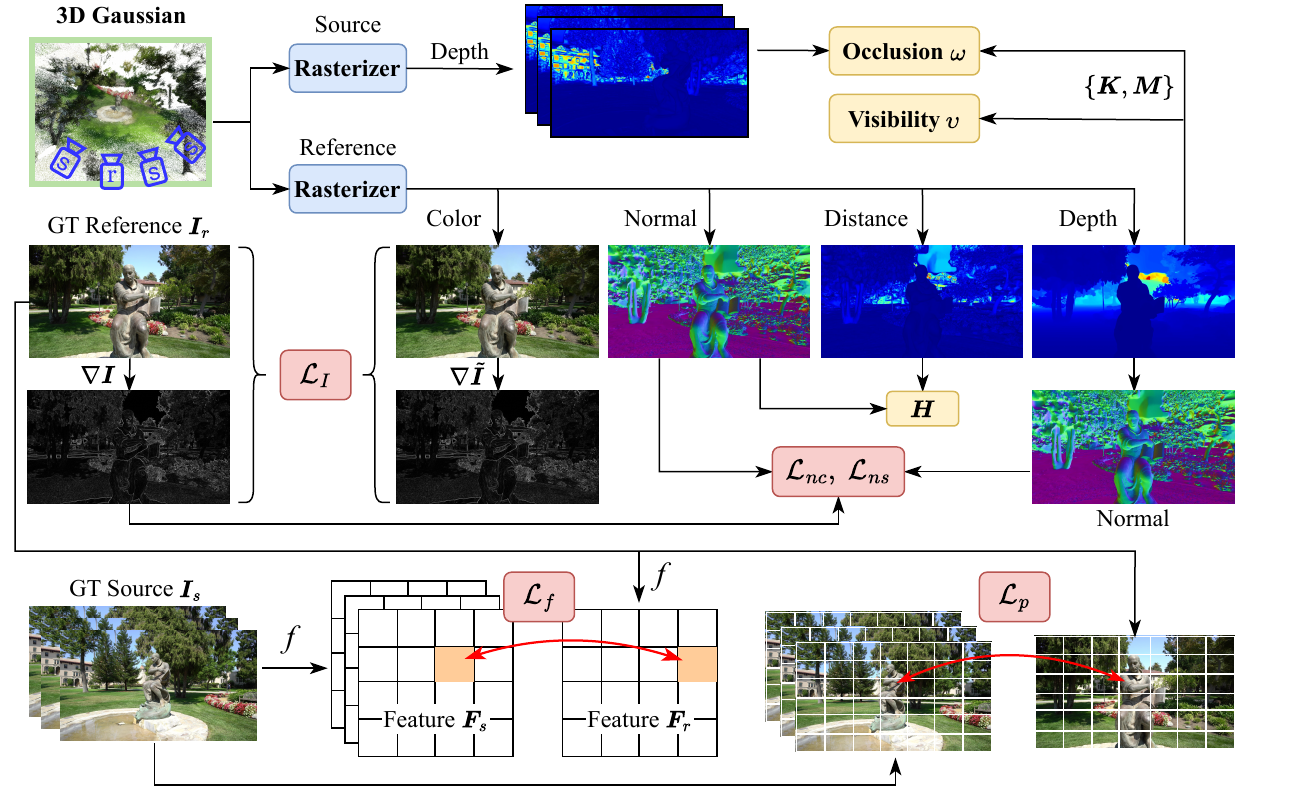}  \vspace{-0.2cm}
    \caption{
        Overview of our method.
        The training includes five loss functions: $\mathcal{L}_{I}$, $\mathcal{L}_{nc}$, $\mathcal{L}_{ns}$, $\mathcal{L}_{p}$ and $\mathcal{L}_{f}$.
        The occlusion weight $\omega$, visibility item $\upsilon$ and homography matrix $\bm{H}$ are involved in $\mathcal{L}_{p}$ and $\mathcal{L}_{f}$.
        The image features $\bm{F}_s$ and $\bm{F}_r$ are extracted using a pretrained network $f$.
        $\{\bm{K}, \bm{M}\}$ is the intrinsic/extrinsic parameter matrix of the camera view.
    }
    \label{fig:net}
    \vspace{-0.3cm}
\end{figure}

%%%%%%%%%%%%%%%%%%%%%%%%%%%%%%%%%%%%%%%%%%%%%%%%%%%%%%%%%%%%
\section{Method}

Fig.~\ref{fig:net} illustrates the overall framework of our approach.
Given a set of posed RGB images, our goal is to learn a bunch of 3D Gaussian functions with associated attributes, such as color, opacity, position and shape, to represent the geometry of a 3D scene.
We introduce novel constraints to enable accurate surface reconstruction while preserving high-quality novel view synthesis.

\subsection{Single-View Alignment}

\textbf{Edge-aware Image Reconstruction.}
% \textbf{Edge-aware Image Alignment.}
The original 3DGS~\cite{kerbl20233d} and its variants typically employ a color rendering loss, which combines the L1 reconstruction error with a D-SSIM term.
While effective for overall image quality, this loss alone is insufficient for accurately capturing object boundaries during surface reconstruction, and it tends to overly smooth high-frequency regions and complex structures.
To address this limitation, we propose an edge-aware image reconstruction loss that encourages the model to better preserve sharp structures and boundary details:
\begin{equation}
    \mathcal{L}_{I} = (1 - \beta_1) \bm{L}_1(\tilde{\bm{I}} - \bm{I}) + \beta_1 \bm{L}_{SSIM}(\tilde{\bm{I}} - \bm{I}) + \beta_2 \bm{L}_1(\nabla \tilde{\bm{I}} - \nabla \bm{I}),
\end{equation}
where $\tilde{\bm{I}}$ is the rendered image, $\bm{I}$ is the ground-truth image, and $\nabla \bm{I}$ denotes the image gradient normalized to the range $[0, 1]$.
$\beta_1$ and $\beta_2$ are weight factors.
The incorporation of gradient-based supervision leads to better preservation of object contours and improves reconstruction quality in boundary and texture-rich regions.

\textbf{Normal-based Geometry Alignment.}
2DGS~\cite{huang20242d} introduces a normal consistency loss that aligns the normals of Gaussian primitives with those derived from the rendered depth map, ensuring that each 2D splat locally approximates the underlying object surface.
However, in boundary regions, the Gaussian primitives often exhibit ambiguous normal directions due to insufficient local support, which can lead to inaccurate geometry reconstruction across different surfaces.
To address this issue, we utilize image edges as proxies for geometric edges, assuming that areas with strong image gradients are likely to correspond to surface discontinuities.
Thus, we adopt an edge-aware normal consistency loss defined as:
\begin{equation}
    \mathcal{L}_{nc} = \frac{1}{\mathcal{I}} \sum_{\bm{p} \in \mathcal{I}} \delta ~\cdot \big\| ~ \hat{\bm{N}} - \tilde{\bm{N}} ~\big\|_1 ~,
\end{equation}
where $\delta = (1 - \nabla \bm{I})^2$ serves as a per-pixel weight~\cite{chen2024pgsr} that downweights loss contributions from edge regions, and $\mathcal{I}$ denotes the set of image pixels.
$\tilde{\bm{N}}$ is the rendered normal, and $\hat{\bm{N}}$ is the normal estimated from the depth map gradient~\cite{huang20242d}.
To compute normal $\hat{\bm{N}}$, we first project four neighboring depth samples into 3D points in the camera coordinate system.
We then estimate the surface normal at pixel $\bm{p}$ by computing the cross product of vectors formed from these projected points, effectively fitting a local plane.

While the above loss enforces the global alignment of Gaussian primitives with the actual surface, noisy primitives can still appear in flat or texture-less regions, leading to abrupt and unnatural changes in surface normals.
Moreover, illumination changes, such as shadows shown in Fig.~\ref{fig:intro}, may introduce false edges during reconstruction.
To address these, we use a normal smoothing loss that encourages local continuity of surface normals by penalizing large discrepancies between adjacent pixels:
\begin{equation}
    \mathcal{L}_{ns} = \frac{1}{\mathcal{I}} \sum_{i,j,k} \delta_k \cdot \mathcal{R} \left( \big| \hat{\bm{N}}_k - \hat{\bm{N}}_{(i,j)} \big| - \tau^2 \right)
    \cdot \left[ \big| \tilde{\bm{N}}_k - \tilde{\bm{N}}_{(i,j)} \big| - \tau \right],
\end{equation}
where $\hat{\bm{N}}_{(i,j)}$ and $\tilde{\bm{N}}_{(i,j)}$ denote the normals at pixel location $(i, j)$, and $k \in \{(i+1, j), (i, j+1)\}$ refers to its neighboring pixels in the horizontal and vertical directions.
$\mathcal{R}(\cdot)$ is the ReLU function, and $[\cdot]$ denotes the Iverson bracket, which evaluates to 1 if the condition inside is true and 0 otherwise.
The threshold $\tau$ and weight $\delta$ help distinguish surface edges and prevents over-smoothing in high-frequency regions.
This loss promotes smoother local geometry while preserving meaningful structural edges, thereby improving the overall surface fidelity.

\subsection{Multi-View Alignment}

\textbf{Multi-View Photometric Alignment.}
While image reconstruction and geometry alignment losses help reduce artifacts and preserve coarse geometry, they often fail to capture fine details.
To address this, we draw inspiration from traditional multi-view stereo (MVS) methods~\cite{schonberger2016structure,campbell2008using,fu2022geo}, which refine surfaces by enforcing photometric consistency across views.
Specifically, they project 3D points derived from depth maps onto multiple views and compare their colors to evaluate consistency.
By introducing a photometric consistency loss based on plane patches, we leverage multi-view observations to resolve geometric ambiguities, particularly at object boundaries, and enhance reconstruction accuracy.

As shown in Fig.~\ref{fig:net}, let $\bm{I}_r$ be the reference view image, and $\bm{I}_s \in \{\bm{I}_{s,i} ~|~ i = 1,2,\ldots,N\}$ denote its neighboring source views.
For a pixel $\bm{p}_r$ in the reference view, we define its corresponding plane by normal $\bm{n}_r$ and distance $\bm{d}_r$.
Using a homography matrix $\bm{H}_{rs}$, $\bm{p}_r$ is projected to $\bm{p}_s^r$ in the source view as follows:
\begin{equation}
    \bm{p}_s^r = \bm{H}_{rs} ~ \bm{p}_r,
    ~~\bm{H}_{rs} = \bm{K}_s \left( \bm{R}_{rs} - \frac{\bm{T}_{rs} \bm{n}_r^\top} {\bm{d}_r} \right) \bm{K}_r^{-1} ~~,
\end{equation}
where $\bm{R}_{rs}$ and $\bm{T}_{rs}$ are the relative rotation and translation from the reference to the source view.
Assuming local planarity, we warp a reference patch $\mathcal{P}_r$ centered at $\bm{p}_r$ to its corresponding source patch $\mathcal{P}_s$ using $\bm{H}_{rs}$.
We enforce multi-view photometric alignment by encouraging consistency between $\mathcal{P}_r$ and $\mathcal{P}_s$:
\begin{equation}  \label{eq:ncc}
    \mathcal{L}_{p} = \sum_{\bm{I}_{s} \in \{\bm{I}_{s,i}\}} \frac{1}{V} \sum_{\bm{p}_r \in \bm{I}_r}
    \upsilon_{rs}(\bm{p}_r) \cdot \omega (\bm{p}_r)
    \cdot \left(1 - \mathcal{C}\big( \mathcal{P}_r(\bm{p}_r), ~\mathcal{P}_s(\bm{p}_s^r) \big) \right), ~~i=1,2,\ldots,N ~,
\end{equation}
where $\mathcal{C}(\cdot)$ is the normalized cross-correlation~\cite{yoo2009fast}, and $V$ is the number of visible pixels.
The visibility term $\upsilon_{rs}(\bm{p}_r)$ indicates whether $\bm{p}_r$ is visible in the source view, and $\omega(\bm{p}_r)$ is a weight accounting for geometric occlusion.
Note that we aggregate the losses from all source views by summation, not averaging.
The definitions of $\upsilon_{rs}(\bm{p}_r)$ and $\omega(\bm{p}_r)$ are detailed in the following.

$\bullet$ Due to viewpoint changes, a 2D pixel $\bm{p}_r$ in the reference view may fall outside the field of view when projected into a source view.
We define a visibility term $\upsilon_{rs}(\bm{p}_r)$ to indicate whether $\bm{p}_r$ is visible from the source viewpoint.
Given a pixel $\bm{p}_r$ with rendered depth $\bm{z}_r$, its corresponding 3D point $\bm{x}_{r}$ and projected pixel coordinate $\bm{p}_{s}'$ in the source view are computed as:
\begin{equation}
    \bm{p}_{s}' = \pi(\bm{K} \bm{M}_{s} \bm{M}_{r}^{-1} \bm{x}_{r}), ~~\bm{x}_{r} = \bm{z}_{r} \bm{K}^{-1} \bar{\bm{p}}_{r} ~,
\end{equation}
where $\bm{M}$ is the extrinsic matrix of the camera, $\pi(\cdot)$ converts 3D coordinates to 2D pixels.
The pixel $\bm{p}_r$ is considered visible in the source view if its projection $\bm{p}_s'$ lies within the image bounds.
Thus the visibility term is defined as:
\begin{equation}
    \upsilon_{rs}(\bm{p}_r) = \big[ (0, 0) < \bm{p}_{s}' < (W, H) \big],
\end{equation}
where $(W, H)$ is the image resolution, and $[\cdot]$ denotes the Iverson bracket.

$\bullet$ During projection via the homography matrix, some pixels may be occluded or exhibit significant geometric error~\cite{chen2024pgsr}.
To avoid the influence of such outliers, we exclude them from the multi-view alignment loss using an occlusion-aware weight.
Given a reference 3D point $\bm{x}_r$ and its corresponding rendered (or interpolated) depth $\bm{z}_s$ in the source view, we first compute the projection error at $\bm{p}_r$ as:
\begin{align}
    \varphi(\bm{p}_r) &= ||~ \bm{p}_r - \bm{p}_r' ~||_2 ~,  \\
    \bm{p}_r' = \pi(\bm{K} \bm{M}_{r} \bm{M}_{s}^{-1} \bm{x}_{s}), ~~& \bm{x}_{s} = \ddot{\bm{x}}_{s}' \cdot \bm{z}_{s}, ~~ \bm{x}_{s}' = \bm{M}_{s} \bm{M}_{r}^{-1} \bm{x}_{r},
\end{align}
where $\bm{p}_r'$ is the reprojected pixel in the reference view, $\ddot{\bm{x}}_{s}'$ denotes the depth normalized version of $\bm{x}_{s}'$.
We then define the occlusion weight as $\omega(\bm{p}_r) = 1/{\rm exp}(\varphi(\bm{p}_r))$ if $\varphi(\bm{p}_r) < 1$, and otherwise 0.
A small projection error indicates reliable geometry, resulting in a higher weight, while a large error implies occlusion or misalignment, thus being downweighted or discarded.

\textbf{Multi-View Feature Alignment.}
The previously introduced image reconstruction and photometric alignment losses help preserve the shape and structure of the objects.
However, image-based losses are susceptible to noise, blur, and low-texture regions.
Additionally, due to lighting variations, the color of the same surface point may differ across views, making photometric consistency unreliable.
To address these limitations, we introduce a multi-view feature alignment loss.
We extract image features using a pretrained network $f$~\cite{zhang2023vis}, \ie, $\bm{F} = f(\bm{I})$.
Let $\bm{F}_r$ denote the reference view's feature map, and $\bm{F}_s$ be one of the source view features, with $\bm{F}_s \in \{\bm{F}_{s,i} ~|~ i = 1,2,\ldots,N \}$.
Then the pixel-wise feature alignment loss is defined as:
\begin{equation}   \label{eqn:feature_loss}
    \mathcal{L}_{f} = \frac{1}{N} \sum_{\bm{F}_{s} \in \{\bm{F}_{s,i}\}} \frac{1}{V} \sum_{\bm{p}_r \in \bm{I}_r}
    \upsilon_{rs}(\bm{p}_r) \cdot \omega (\bm{p}_r)
    \cdot \left|1 - \cos \big( \bm{F}_{r}(\bm{p}_{r}), ~\bm{F}_{s}(\bm{p}_{s}') \big)\right|, ~~i=1,2,\ldots,N.
\end{equation}
where $\cos(\cdot)$ denotes the cosine similarity between feature vectors.
% The weights $\upsilon_{rs}(\bm{p}_r)$ and $\omega(\bm{p}_r)$ serve the same purposes as in the photometric alignment loss, accounting for visibility and occlusion, respectively.
This feature-level loss improves robustness under challenging conditions such as appearance variation and poor lighting consistency.

\textbf{Final loss.}
To summarize, the final training objective integrates five components:
\begin{equation}
    \mathcal{L} = \mathcal{L}_{I} + \lambda_1 \mathcal{L}_{nc} + \lambda_2 \mathcal{L}_{ns} + \lambda_3 \mathcal{L}_{p} + \lambda_4 \mathcal{L}_{f} ~,
\end{equation}
where $\lambda_1$, $\lambda_2$, $\lambda_3$ and $\lambda_4$ are weighting factors determined based on validation performance.

\begin{figure}[t]
    \centering
    \includegraphics[width=\linewidth]{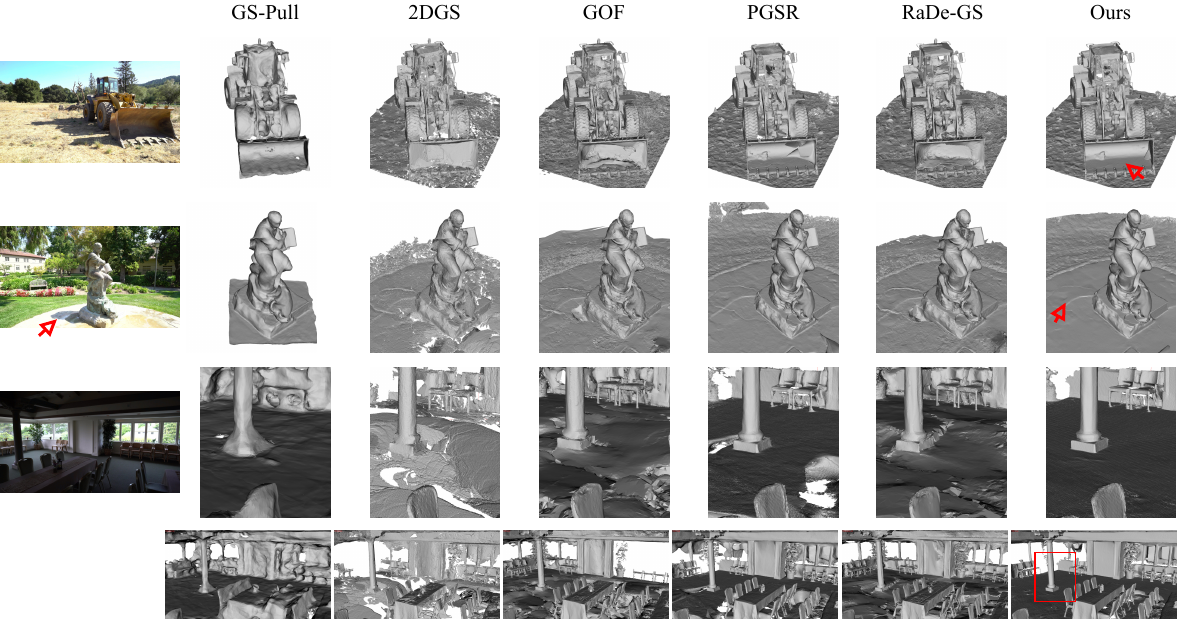}  \vspace{-0.6cm}
    \caption{
        Visual comparison of surface reconstruction results on the TNT dataset.
        Our method can handle shadows and large indoor flat regions.
        GS-Pull reconstructs only the foreground objects.
    }
    \label{fig:tnt}
    \vspace{-0.2cm}
\end{figure}

\begin{figure}[t]
    \centering
    \includegraphics[width=\linewidth]{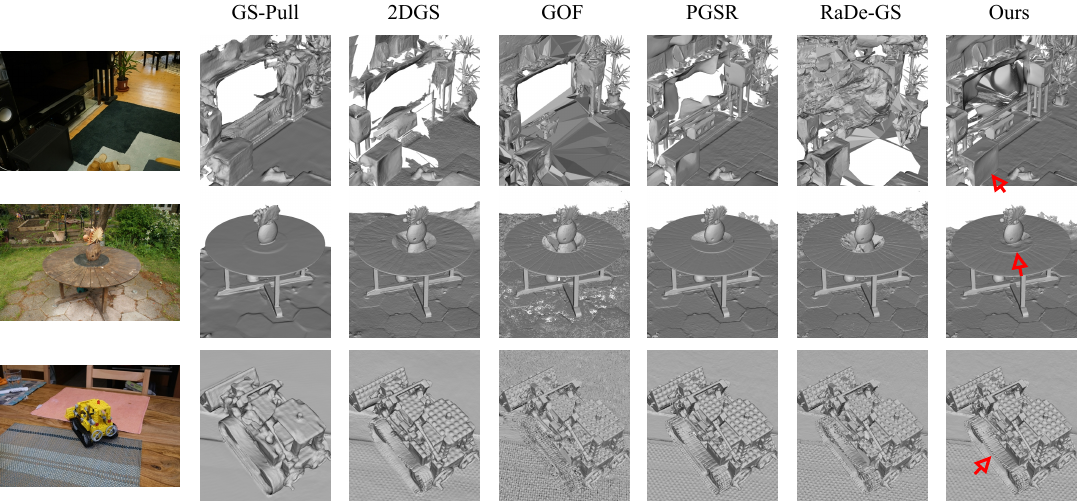}  \vspace{-0.7cm}
    \caption{
        Visual comparison of surface reconstruction results on the Mip-NeRF 360 dataset.
        Our approach effectively handles the challenges posed by cluttered lighting and boundaries.
    }
    \label{fig:360}
    \vspace{-0.3cm}
\end{figure}

\begin{table*}[t]
\centering
% \small
\footnotesize
\setlength{\tabcolsep}{0.8mm}
\caption{
    Quantitative comparison of Chamfer distances on the DTU dataset.
    The best results are highlighted as \colorbox{orange}{1st}, \colorbox{pink}{2nd} and \colorbox{yellow}{3rd}.
    $\ast$ means that the source code is not available.
}
% \vspace{-0.2cm}
\resizebox{\textwidth}{!}{
\begin{tabular}{@{}llcccccccccccccccccc}
    \toprule
    \multicolumn{2}{c}{} & 24 & 37 & 40 & 55 & 63 & 65 & 69 & 83 & 97 & 105 & 106 & 110 & 114 & 118 & 122 && \textbf{Mean} & Time \\

    \cmidrule(r){3-17} \cmidrule(l){19-20}
    \multirow{6}{*}{\rotatebox[origin=c]{90}{Implicit}}

    & NeRF~\cite{mildenhall2020nerf}         & 1.90        & 1.60        & 1.85        & 0.58        & 2.28        & 1.27        & 1.47        & 1.67        & 2.05        & 1.07        & 0.88        & 2.53        & 1.06        & 1.15        & 0.96        && 1.49        & $>$12h          \\
    & VolSDF~\cite{yariv2021volume}          & 1.14        & 1.26        & 0.81        & 0.49        & 1.25        & 0.70        & 0.72        & 1.29        & 1.18        & 0.70        & 0.66        & 1.08        & 0.42        & 0.61        & 0.55        && 0.86        & $>$12h          \\
    & NeuS~\cite{wang2021neus}               & 1.00        & 1.37        & 0.93        & 0.43        & 1.10        & 0.65        & 0.57        & 1.48        & 1.09        & 0.83        & 0.52        & 1.20        & 0.35        & 0.49        & 0.54        && 0.84        & $>$12h          \\
    & NeuralWarp~\cite{darmon2022improving}  & 0.49        & 0.71        & 0.38        & 0.38        & 0.79        & 0.81        & 0.82        & 1.20        & 1.06        & 0.68        & 0.66        & 0.74        & 0.41        & 0.63        & 0.51        && 0.68        & $>$10h          \\
    & Neuralangelo~\cite{li2023neuralangelo} & 0.37        & 0.72        & 0.35        & 0.35        & 0.87        & \best0.54   & 0.53        & 1.29        & 0.97        & 0.73        & \tbest0.47  & 0.74        & 0.32        & 0.41        & 0.43        && 0.61        & $>$12h          \\
    & PSDF$^\ast$~\cite{su2024psdf}          & 0.36        & 0.60        & 0.35        & 0.36        & \best0.70   & \tbest0.61  & \sbest0.49  & 1.11        & 0.89        & \tbest0.60  & \tbest0.47  & \tbest0.57  & \sbest0.30  & \tbest0.40  & \sbest0.37  && 0.55        & -               \\

    \cmidrule(r){1-2} \cmidrule(r){3-17} \cmidrule(l){19-20}
    \multirow{10}{*}{\rotatebox[origin=c]{90}{Explicit}}

    & 3DGS~\cite{kerbl20233d}                & 2.14        & 1.53        & 2.08        & 1.68        & 3.49        & 2.21        & 1.43        & 2.07        & 2.22        & 1.75        & 1.79        & 2.55        & 1.53        & 1.52        & 1.50        && 1.96        & \best3.4m       \\
    & SuGaR~\cite{guedon2024sugar}           & 1.47        & 1.33        & 1.13        & 0.61        & 2.25        & 1.71        & 1.15        & 1.63        & 1.62        & 1.07        & 0.79        & 2.45        & 0.98        & 0.88        & 0.79        && 1.33        & 1h              \\%
    & GaussianSurfels\cite{dai2024high}      & 0.66        & 0.93        & 0.54        & 0.41        & 1.06        & 1.14        & 0.85        & 1.29        & 1.53        & 0.79        & 0.82        & 1.58        & 0.45        & 0.66        & 0.53        && 0.88        & \sbest4.5m      \\
    & 2DGS~\cite{huang20242d}                & 0.48        & 0.91        & 0.39        & 0.39        & 1.01        & 0.83        & 0.81        & 1.36        & 1.27        & 0.76        & 0.70        & 1.40        & 0.40        & 0.76        & 0.52        && 0.80        & 5.8m            \\
    & GS-Pull~\cite{zhang2024neural}         & 0.51        & \tbest0.56  & 0.46        & 0.39        & 0.82        & 0.67        & 0.85        & 1.37        & 1.25        & 0.73        & 0.54        & 1.39        & 0.35        & 0.88        & 0.42        && 0.75        & \tbest5.6m      \\
    & GOF~\cite{yu2024gaussian}              & 0.50        & 0.82        & 0.37        & 0.37        & 1.12        & 0.74        & 0.73        & 1.18        & 1.29        & 0.68        & 0.77        & 0.90        & 0.42        & 0.66        & 0.49        && 0.74        & 32m             \\
    & RaDe-GS~\cite{zhang2024rade}           & 0.46        & 0.73        & \tbest0.33  & 0.38        & 0.79        & 0.75        & 0.76        & 1.19        & 1.22        & 0.62        & 0.70        & 0.78        & 0.36        & 0.68        & 0.47        && 0.68        & 6.5m            \\
    & PGSR~\cite{chen2024pgsr}               & \sbest0.34  & 0.58        & \best0.29   & \best0.29   & \tbest0.78  & \sbest0.58  & 0.54        & \best1.01   & \tbest0.73  & \best0.51   & 0.49        & 0.69        & \tbest0.31  & \sbest0.37  & \tbest0.38  && \tbest0.53  & 15m             \\
    & GausSurf$^\ast$~\cite{wang2024gaussurf}& \tbest0.35  & \sbest0.55  & 0.34        & \tbest0.34  & \sbest0.77  & \sbest0.58  & \tbest0.51  & \tbest1.10  & \sbest0.69  & \tbest0.60  & \sbest0.43  & \best0.49   & 0.32        & \tbest0.40  & \sbest0.37  && \sbest0.52  & -               \\
    & Ours                                   & \best0.32   & \best0.49   & \sbest0.32  & \sbest0.30  & \sbest0.77  & 0.68        & \best0.43   & \sbest1.05  & \best0.61   & \sbest0.57  & \best0.36   & \sbest0.52  & \best0.28   & \best0.33   & \best0.30   && \best0.49   & 15.5m           \\
    \bottomrule
\end{tabular} }
\label{tab:dtu}
\vspace{-0.3cm}
\end{table*}
\begin{table}[t]
\centering
% \small
\footnotesize
\setlength{\tabcolsep}{1.2mm}
\caption{
    Quantitative comparison of F1-scores on the TNT dataset.
    The best results are highlighted as \colorbox{orange}{1st}, \colorbox{pink}{2nd} and \colorbox{yellow}{3rd}.
    $\ast$ means that the source code is not available.
}
\vspace{-0.2cm}
\label{tab:tnt}
% \resizebox{\textwidth}{!}{
\begin{tabular}{@{}llcccccclcc}
    \toprule
    \multicolumn{2}{c}{} & Barn & Caterpillar & Courthouse & Ignatius & Meetingroom & Truck && \textbf{Mean} & Time  \\

    \cmidrule(){3-8} \cmidrule(l){10-11}
    \multirow{4}{*}{\rotatebox[origin=c]{90}{Implicit}}

    & NeuS~\cite{wang2021neus}               & 0.29       & 0.29       & 0.17       & \sbest0.83 & 0.24       & 0.45            && 0.38       & $>$12h       \\
    & Geo-Neus~\cite{fu2022geo}              & 0.33       & 0.26       & 0.12       & 0.72       & 0.20       & 0.45            && 0.35       & $>$12h       \\
    & Neuralangelo~\cite{li2023neuralangelo} & \sbest0.70 & 0.36       & \tbest0.28 & \best0.89  & 0.32       & 0.48            && 0.50       & $>$12h       \\
    & PSDF$^\ast$~\cite{su2024psdf}          & 0.62       & 0.39       & \best0.42  & 0.79       & \best0.47  & 0.53            && \sbest0.53 & -            \\

    \cmidrule(r){1-2} \cmidrule(){3-8} \cmidrule(l){10-11}
    \multirow{11}{*}{\rotatebox[origin=c]{90}{Explicit}}

    & 3DGS~\cite{kerbl20233d}                & 0.13       & 0.08       & 0.09       & 0.04       & 0.01       & 0.19            && 0.09       & \sbest7.5m  \\
    & DN-Splatter~\cite{turkulainen2025dn}   & 0.15       & 0.11       & 0.07       & 0.18       & 0.01       & 0.20            && 0.12       & 20m         \\%
    & SuGaR~\cite{guedon2024sugar}           & 0.14       & 0.16       & 0.08       & 0.33       & 0.15       & 0.26            && 0.19       & 2h          \\%
    & GaussianSurfels~\cite{dai2024high}     & 0.24       & 0.22       & 0.07       & 0.39       & 0.12       & 0.24            && 0.21       & \best5m     \\
    & 2DGS~\cite{huang20242d}                & 0.41       & 0.23       & 0.16       & 0.51       & 0.17       & 0.45            && 0.32       & \sbest7.5m  \\
    & GS-Pull~\cite{zhang2024neural}         & 0.60       & 0.37       & 0.16       & 0.71       & 0.22       & 0.52            && 0.43       & 18m         \\%
    & GOF~\cite{yu2024gaussian}              & 0.51       & 0.41       & \tbest0.28 & 0.68       & 0.28       & 0.59            && 0.46       & 40m         \\
    & RaDe-GS~\cite{zhang2024rade}           & 0.43       & 0.32       & 0.21       & 0.69       & 0.25       & 0.51            && 0.40       & \tbest9m    \\
    & PGSR~\cite{chen2024pgsr}               & \tbest0.66 & \sbest0.44 & 0.20       & 0.81       & 0.33       & \best0.66       && \tbest0.52 & 25.5m       \\
    & GausSurf$^\ast$~\cite{wang2024gaussurf}& 0.50       & \tbest0.42 & \sbest0.30 & 0.73       & \tbest0.39 & \sbest0.65      && 0.50       & -           \\
    & Ours                                   & \best0.71  &\best0.45   & 0.21       & \tbest0.82 & \sbest0.40 & \tbest0.64      && \best0.54  & 20.6m       \\

    \bottomrule
\end{tabular} %}
\vspace{-0.4cm}
\end{table}
\begin{table}[t]
\centering
% \small
\footnotesize
\setlength{\tabcolsep}{1.1mm}
\caption{
    Quantitative comparison on the Mip-NeRF 360 dataset.
    The best results are highlighted as \colorbox{orange}{1st}, \colorbox{pink}{2nd} and \colorbox{yellow}{3rd}.
}
\vspace{-0.2cm}
\label{tab:mipnerf360}
% \resizebox{\textwidth}{!}{
\begin{tabular}{@{}lccccccccc}
    \toprule
    & \multicolumn{3}{c}{\textbf{Outdoor scenes}}               & \multicolumn{3}{c}{\textbf{Indoor scenes}}                & \multicolumn{3}{c}{\textbf{Average on all scenes}}             \\
    & PSNR~$\uparrow$ & SSIM~$\uparrow$ & LPIPS~$\downarrow$    & PSNR~$\uparrow$ & SSIM~$\uparrow$ & LPIPS~$\downarrow$    & PSNR~$\uparrow$ & SSIM~$\uparrow$ & LPIPS~$\downarrow$         \\
    \cmidrule(lr){2-4} \cmidrule(l){5-7} \cmidrule(l){8-10}

    NeRF~\cite{mildenhall2020nerf}         & 21.46       & 0.458       & 0.515          & 26.84        & 0.790       & 0.370        & 23.85        & 0.606         & 0.451        \\
    Deep Blending~\cite{hedman2018deep}    & 21.54       & 0.524       & 0.364          & 26.40        & 0.844       & 0.261        & 23.70        & 0.666         & 0.318        \\
    Instant NGP~\cite{muller2022instant}   & 22.90       & 0.566       & 0.371          & 29.15        & 0.880       & 0.216        & 25.68        & 0.706         & 0.302        \\
    MERF~\cite{reiser2023merf}             & 23.19       & 0.616       & 0.343          & 27.80        & 0.855       & 0.271        & 25.24        & 0.722         & 0.311        \\
    BakedSDF~\cite{yariv2023bakedsdf}      & 22.47       & 0.585       & 0.349          & 27.06        & 0.836       & 0.258        & 24.51        & 0.697         & 0.309        \\
    Mip-NeRF 360~\cite{barron2022mip}      & 24.47       & 0.691       & 0.283          & \best31.72   & 0.917       & 0.180        & \best27.69   & 0.791         & 0.237        \\

    \cmidrule(r){1-1} \cmidrule(lr){2-4} \cmidrule(l){5-7} \cmidrule(l){8-10}

    3DGS~\cite{kerbl20233d}                & 24.64       & 0.731       & 0.234           & 30.41       & 0.920       & 0.189        & 27.20        & 0.815         & 0.214         \\
    SuGaR~\cite{guedon2024sugar}           & 22.93       & 0.629       & 0.356           & 29.43       & 0.906       & 0.225        & 25.82        & 0.752         & 0.298         \\
    2DGS~\cite{huang20242d}                & 24.34       & 0.717       & 0.246           & 30.40       & 0.916       & 0.195        & 27.03        & 0.805         & 0.223         \\
    GS-Pull~\cite{zhang2024neural}         & 23.76       & 0.703       & 0.278           & \tbest30.78 & 0.925       & 0.182        & 26.88        & 0.802         & 0.235         \\
    GOF~\cite{yu2024gaussian}              & 24.82       & 0.750       & \tbest0.202     & \sbest30.79 & 0.924       & 0.184        & 27.47        & \sbest0.827   & \tbest0.194   \\
    RaDe-GS~\cite{zhang2024rade}           & \best25.17  & \best0.764  & \sbest0.199     & 30.74       & \tbest0.928 & \tbest0.165  & \sbest27.65  & \best0.837    & \sbest0.184   \\
    PGSR~\cite{chen2024pgsr}               & 24.45       & 0.730       & 0.224           & 30.41       & \sbest0.930 & \sbest0.161  & 27.10        & \tbest0.819   & 0.196         \\
    GausSurf~\cite{wang2024gaussurf}       & \sbest25.09 & \tbest0.753 & 0.212           & 30.05       & 0.920       & 0.183        & 27.29        & \sbest0.827   & 0.199         \\
    % Ours                                   & \tbest24.97 & \sbest0.758 & \best0.192      & 30.64       & \best0.933  & \best0.153   & \tbest27.49  & \best0.836    & \best0.175    \\
    Ours                                   & \tbest25.00 & \sbest0.760 & \best0.191      & 30.63       & \best0.933  & \best0.153   & \tbest27.50  & \best0.837    & \best0.174    \\
    \bottomrule
\end{tabular} %}
\vspace{-0.5cm}
\end{table}

%%%%%%%%%%%%%%%%%%%%%%%%%%%%%%%%%%%%%%%%%%%%%%%%%%%%%%%%%%%%
\section{Experiments}

\textbf{Evaluation Protocols.}
We evaluate our surface reconstruction performance on the DTU~\cite{jensen2014large} and Tanks and Temples (TNT)~\cite{knapitsch2017tanks} datasets.
Following prior works~\cite{huang20242d,yu2024gaussian,chen2024pgsr,zhang2024rade}, we use 15 scenes from the DTU dataset and 6 scenes from the TNT dataset for evaluation.
Depth maps are rendered for all training views, and a TSDF~\cite{curless1996volumetric} is constructed for mesh extraction.
For novel view synthesis, we use the Mip-NeRF 360 dataset~\cite{barron2022mip}, which contains large-scale indoor and outdoor scenes with complex lighting and fine-grained geometric details.
Following 3DGS~\cite{kerbl20233d}, one out of every eight images is used for evaluation, while the remaining seven are used for training.
We employ COLMAP~\cite{schoenberger2016sfm} to generate a sparse point cloud from the original dataset images for initializing the 3D Gaussians.
All images are downsampled to a lower resolution to facilitate training.
Following established protocols~\cite{huang20242d,yu2024gaussian,chen2024pgsr,zhang2024rade}, we report Chamfer distance for surface reconstruction on the DTU dataset and F1-score for the TNT dataset.
For novel view synthesis, we evaluate using three widely adopted image quality metrics: PSNR, SSIM, and LPIPS.

\textbf{Implementation Details.}
Our overall pipeline, training strategy, and hyperparameter settings generally follow 3DGS~\cite{kerbl20233d}.
We set the number of source views to $N \!=\! 3$, the threshold in $\mathcal{L}_{ns}$ to $\tau \!=\! 0.01$, and the patch size in $\mathcal{L}_{p}$ to $7 \!\times\! 7$.
The loss weight factors are set as follows: $\beta_1 \!=\! 0.2$, $\beta_2 \!=\! 0.03$, $\lambda_1 \!=\! 0.015$, $\lambda_2 \!=\! 0.3$, $\lambda_3 \!=\! 0.15$, and $\lambda_4 \!=\! 1.0$.
The model is trained for 20,000 iterations for surface reconstruction and 30,000 iterations for novel view synthesis.
We first pretrain the model using only the color loss for 7,000 steps to obtain a coarse geometric initialization, which provides a stable foundation for subsequent geometry refinement.
Then, we incorporate our image edge item and normal-based geometry alignment into the training.
To further refine geometry, we sequentially apply our multi-view photometric alignment for 8,000 iterations, followed by 5,000 iterations of multi-view feature alignment.
For novel view synthesis, we continue training for an additional 10,000 steps to optimize rendering quality.
All experiments are conducted on a single NVIDIA RTX 4090 GPU.

\subsection{Performance Evaluation}

\textbf{Comparisons on DTU.}
We first compare our method with state-of-the-art implicit and explicit surface reconstruction approaches on the DTU dataset~\cite{jensen2014large}.
Following standard protocol, reconstructions are clipped using the provided mask, and evaluations are performed only on foreground objects, as the ground truth point clouds exclude background regions.
As shown in Table~\ref{tab:dtu}, our method achieves the lowest average Chamfer distance and ranks best across most scenes.
Compared to implicit approaches such as NeuS~\cite{wang2021neus} and Neuralangelo~\cite{li2023neuralangelo}, our method delivers significantly better reconstruction accuracy while being much more efficient in terms of runtime.
It is worth noting that most implicit methods~\cite{wang2021neus,li2023neuralangelo} only reconstruct foreground geometry, whereas our approach can produce detailed and complete meshes, including background regions, which is an essential feature for mesh-based rendering.
Although our method is slightly slower than 3DGS~\cite{kerbl20233d} and 2DGS~\cite{huang20242d} due to the use of multi-view alignment, it achieves significant improvements in reconstruction quality over these earlier Gaussian-based methods.

\textbf{Comparisons on TNT.}
We further evaluate our method on the TNT dataset~\cite{knapitsch2017tanks}, comparing it against both implicit and explicit surface reconstruction baselines.
Since the ground-truth point clouds do not include background regions, the evaluation is restricted to foreground objects.
As shown in Table~\ref{tab:tnt}, our method achieves the best reconstruction performance among all competing approaches, including both implicit and explicit methods.
Notably, while several Gaussian-based methods require less optimization time, they tend to produce results with much lower accuracy.
In contrast, our method reaches a better balance between efficiency and reconstruction quality.
For example, GS-Pull~\cite{zhang2024neural} only reconstructs foreground objects and often generates overly smooth surfaces.
Fig.~\ref{fig:tnt} provides a qualitative comparison.
Our method produces more accurate and detailed reconstructions for both foreground and background regions.
It also effectively mitigates the impact of shadows, whereas baseline methods often yield noisy meshes or fail to capture geometric details.
The use of geometry, photometric, and feature-based alignment from multiple views provides strong guidance, enabling the Gaussian primitives to converge more accurately to the true surface geometry.

\textbf{Comparisons on Mip-NeRF 360.}
We also evaluate our approach on the Mip-NeRF 360 dataset~\cite{barron2022mip} for novel view synthesis.
Table~\ref{tab:mipnerf360} reports quantitative comparisons against state-of-the-art Gaussian-based and other neural rendering baselines.
Our method outperforms competitors on most metrics, demonstrating superior image fitting and generalization to unseen viewpoints.
This evidences that our enhanced geometry representation yields higher visual fidelity.
Notably, the Mip-NeRF 360 itself achieves the highest average PSNR on indoor scenes but lags on SSIM and LPIPS.
Among Gaussian-based methods, 2DGS~\cite{huang20242d}, SuGaR~\cite{guedon2024sugar}, and GS-Pull~\cite{zhang2024neural} perform worse than vanilla 3DGS~\cite{kerbl20233d}, suggesting that their planar Gaussian constraints degrade performance in complex environments.
Our ablation results in Table~\ref{tab:ablation} further confirm that flattening 3D Gaussians into planar Gaussian disks is ineffective for our framework.
Our method preserves the full 3D Gaussian representation and delivers high-quality surfaces without sacrificing novel-view rendering quality.
Fig.~\ref{fig:360} provides a qualitative comparison of reconstructed meshes.
Consistent with our observations on the TNT dataset, our method recovers more accurate and complete surfaces in both foreground and background regions, whereas other methods suffer from noise, oversmoothing, or missing details, especially in challenging indoor scenes.

% \begin{table}[t]
\begin{wraptable}{r}{0.5\textwidth}
\centering
% \small
\footnotesize
\setlength{\tabcolsep}{1.0mm}
\vspace{-1.0cm}
\caption{
    Ablations on the TNT dataset.
}
\vspace{0.1cm}
\label{tab:ablation}
% \resizebox{0.92\columnwidth}{!}{
\begin{tabular}{lccc}
    \toprule
    & Precision~$\uparrow$ & Recall~$\uparrow$ & F1-score~$\uparrow$                                    \\
    \midrule
    Only $\mathcal{L}_I$                    & 0.09               & 0.23               & 0.13            \\
    w/o edge item                           & 0.49               & 0.59               & 0.53            \\
    w/o weight $\delta$                     & 0.50               & 0.59               & 0.53            \\
    w/o $\mathcal{L}_{nc}$                  & 0.48               & \textbf{0.60}      & 0.52            \\
    w/o $\mathcal{L}_{ns}$                  & 0.47               & 0.58               & 0.51            \\
    w/o $\mathcal{L}_{nc}+\mathcal{L}_{ns}$ & 0.40               & 0.57               & 0.46            \\
    w/o $\mathcal{L}_{p}$                   & 0.46               & 0.56               & 0.50            \\
    w/o $\mathcal{L}_{f}$                   & 0.49               & \textbf{0.60}      & 0.53            \\
    w/o $\mathcal{L}_{p}+\mathcal{L}_{f}$   & 0.33               & 0.40               & 0.36            \\
    w/ scale loss                           & \textbf{0.51}      & \textbf{0.60}      & \textbf{0.54}   \\
    $N = 1$                                 & 0.49               & 0.58               & 0.52            \\
    $N = 2$                                 & 0.49               & 0.59               & 0.53            \\
    $N = 4$                                 & \textbf{0.51}      & \textbf{0.60}      & \textbf{0.54}   \\
    \midrule
    Ours                                    & \textbf{0.51}      & \textbf{0.60}      & \textbf{0.54}   \\
    \bottomrule
\end{tabular}  %}
\vspace{-0.3cm}
% \end{table}
\end{wraptable}

\subsection{Ablation Studies}

To quantify the contributions of our alignment constraints, we perform ablations by selectively removing loss terms and report reconstruction quality on the TNT dataset.
In addition to the \textit{F1-score}, we also report \textit{Precision} and \textit{Recall} to provide a more comprehensive evaluation.
The base color rendering loss from 3DGS is always retained in the following experiments.
We provide quantitative results in Table~\ref{tab:ablation}.

\vspace{-0.1cm} (1) \textit{Only image reconstruction loss ($\mathcal{L}_I$)}:
Removing all alignment losses yields the worst results, with an average F1-score of $0.13$, but still better than the vanilla 3DGS's score of $0.09$.

\vspace{-0.1cm} (2) \textit{Edge-aware term in $\mathcal{L}_I$}:
Omitting the image edge-based component slightly degrades performance, confirming its role in preserving boundary detail.

\vspace{-0.1cm} (3) \textit{Edge-aware weight $\delta$}:
In boundary regions, Gaussian primitives often exhibit ambiguous or noisy normal directions, which can lead to incorrect supervision signals.
The weight $\delta$ in loss $\mathcal{L}_{nc}$ reduces the loss contribution from these areas, allowing the network to focus learning on more reliable surface regions.
While the improvement is modest, it reflects the fact that shape boundaries constitute a relatively small proportion of the scene, and thus affect only a small number of sampled points during evaluation.

\vspace{-0.1cm} (4) \textit{Normal-based alignment ($\mathcal{L}_{nc}$, $\mathcal{L}_{ns}$)}:
The normal consistency ($\mathcal{L}_{nc}$) and smoothing ($\mathcal{L}_{ns}$) losses are critical.
Excluding either term causes a noticeable drop in Precision and F1-score, and removing both leads to a dramatic performance collapse.

\vspace{-0.1cm} (5) \textit{Multi-view alignment ($\mathcal{L}_p$, $\mathcal{L}_f$)}:
Enforcing photometric and feature consistency across views consistently improves reconstruction accuracy.
Each multi-view alignment term contributes positively, validating the benefit of cross-view geometric constraints.

\vspace{-0.1cm} (6) \textit{Scale regularization}:
The scaling matrix $\bm{S}$ represents the stretching of a spherical Gaussian along the three axes.
Different from previous works~\cite{chen2024pgsr,cheng2024gaussianpro,zhang2024neural}, incorporating the widely used scale penalty into our method to flatten the 3D Gaussian disks provides no performance gains, and even degrades novel-view rendering quality on the Mip-NeRF 360 dataset.

\vspace{-0.1cm} (7) \textit{Number of source views ($N$)}:
Our method takes both a reference view and $N$ source views.
Increasing the number of source views used in the alignment losses improves reconstruction quality.
However, setting $N=4$ yields no additional performance gains but increases the computational cost.
We therefore choose $N=3$ to balance accuracy and efficiency.

Overall, these ablations demonstrate that each of our proposed alignment constraints plays a distinct and essential role in achieving high-fidelity surface reconstruction.

%%%%%%%%%%%%%%%%%%%%%%%%%%%%%%%%%%%%%%%%%%%%%%%%%%%%%%%%%%%%
\section{Conclusion}

In this paper, we address the limitations of existing 3D Gaussian Splatting approaches in recovering accurate and detailed surface geometry, especially under challenging conditions such as complex lighting and ambiguous object boundaries.
We propose a novel method that improves geometric fidelity by integrating edge-aware supervision, visibility-aware multi-view alignment, and robust geometric constraints based on surface normals and deep visual features.
These components jointly enforce cross-view consistency, enhance boundary sharpness, and mitigate the impact of illumination-induced artifacts.
Extensive experiments demonstrate that our method achieves state-of-the-art performance in both surface reconstruction and novel view synthesis, underscoring its effectiveness and robustness in complex real-world scenarios.
The main limitation of our approach is its relatively slower training speed compared to earlier 3DGS variants.
In future work, we aim to explore adaptive Gaussian pruning and learned covariance regularization to accelerate training and further improve robustness in large-scale and dynamic scenes.

\section*{Acknowledgements}
This work was supported by the National Natural Science Foundation of China (62402401), the Sichuan Provincial Natural Science Foundation of China (2025ZNSFSC1462) and the Fundamental Research Funds for the Central Universities (2682025CX109).

%%% Bibliography
{
\small
\bibliographystyle{ieeenat_fullname}
\bibliography{egbib}
}

% \clearpage

\appendix

This supplementary document is organized as follows:
(1) We first present an overview of the baseline methods used in our evaluation experiments.
(2) We then provide additional qualitative and quantitative experimental results to complement those in the main paper.
(3) Further ablation studies are included to analyze the impact of key components in our method.
(4) More discussion is provided to give a precise understanding of our approach.
(5) We discuss the current limitations of our approach and outline potential directions for future research.
(6) Finally, we consider the potential negative societal impacts associated with our work.

%%%%%%%%%%%%%%%%%%%%%%%%%%%%%%%%%%%%%%%%%%%%%%%%%%%%%%%
\section{Baseline Methods}

We compare our approach with state-of-the-art surface reconstruction and novel view synthesis methods based on 3D Gaussian Splatting, such as 3DGS~\cite{kerbl20233d}, 2DGS~\cite{huang20242d}, GOF~\cite{yu2024gaussian}, and PGSR~\cite{chen2024pgsr}.
In addition, we evaluate against implicit NeRF-based methods, such as NeuS~\cite{wang2021neus} and Neuralangelo~\cite{li2023neuralangelo}, which utilize Signed Distance Functions (SDFs) to represent scenes and convert them into opacity fields for volume rendering via ray tracing.

In our main paper, we report baseline results using values provided in the respective original publications whenever available.
For visual comparisons shown in our figures, we generate the results using the official implementations released by the authors.

%%%%%%%%%%%%%%%%%%%%%%%%%%%%%%%%%%%%%%%%%%%%%%%%%%%%%%%
\section{More Evaluation Results}

\subsection{Comparisons on the Deep Blending Dataset}

In order to further test the performance of the algorithm on different real-world data, we evaluate both novel view synthesis and surface reconstruction on two commonly used indoor scenes, Dr Johnson and Playroom, from the Deep Blending dataset~\cite{hedman2018deep}.
Table~\ref{tab:db} presents quantitative comparisons against state-of-the-art Gaussian-based methods on novel view synthesis.
All results, unless otherwise noted, are obtained from our own re-implementations using the official code released by the respective authors.
For 3DGS~\cite{kerbl20233d}, we use the pretrained models provided by the authors.
For SuGaR~\cite{guedon2024sugar}, we report numbers directly from the original paper to avoid potential inconsistencies.
Our method consistently outperforms all baselines across all metrics, demonstrating superior rendering quality and generalization to unseen viewpoints.
These improvements can be attributed to our enhanced geometric representation, which yields better visual fidelity.
Consistent with observations in the evaluation on the Mip-NeRF 360 dataset~\cite{barron2022mip} presented in the main paper, methods such as 2DGS~\cite{huang20242d}, SuGaR~\cite{guedon2024sugar}, and GS-Pull~\cite{zhang2024neural} underperform compared to vanilla 3DGS~\cite{kerbl20233d}, suggesting that imposing planar Gaussian constraints may hinder performance in complex scenes.

\begin{figure}[t]
    \centering
    \includegraphics[width=\linewidth]{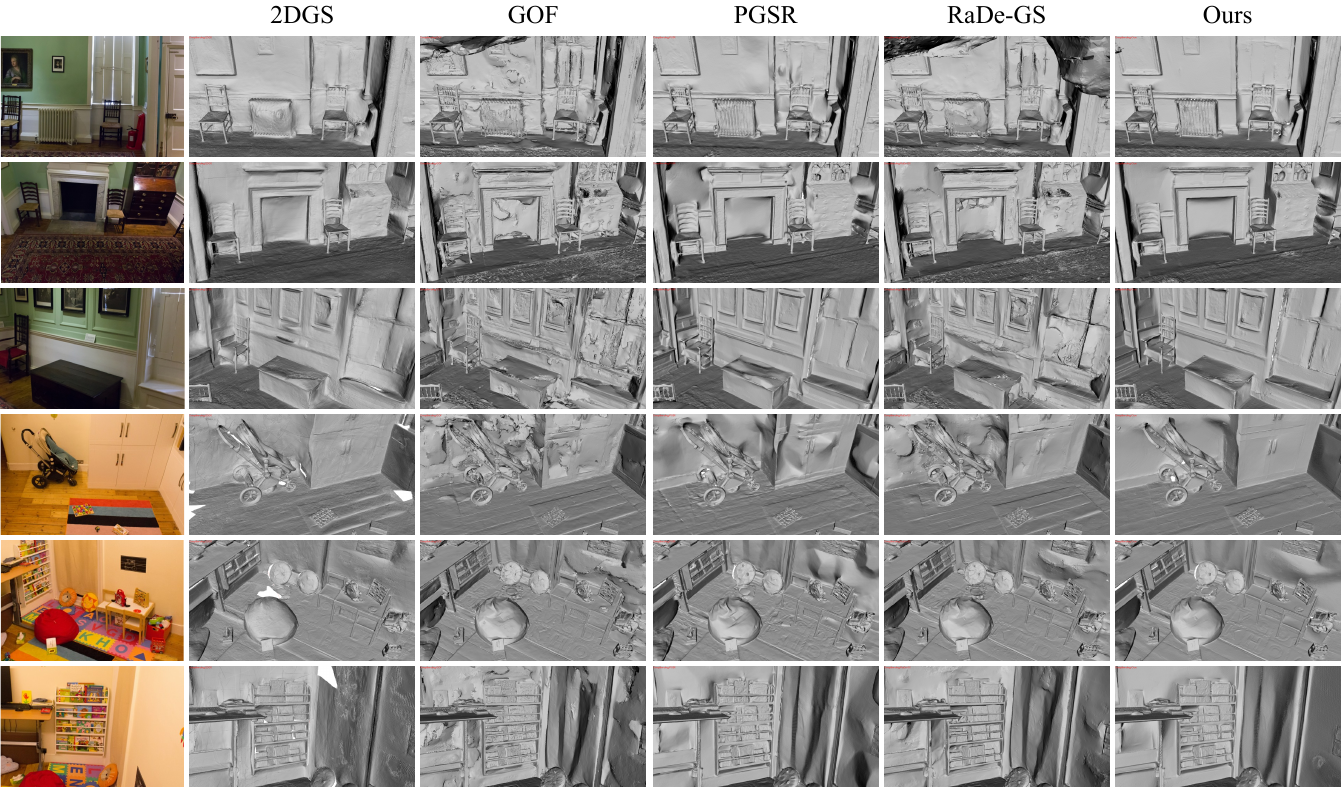}  \vspace{-0.6cm}
    \caption{
        Visual comparison of surface reconstruction results on the Deep Blending dataset.
        Our method effectively handles the challenges posed by complex lighting conditions and ambiguous boundaries.
        GS-Pull is omitted as it fails to produce reasonable reconstructions.
    }
    \label{fig:db}
    % \vspace{-0.1cm}
\end{figure}

Fig.~\ref{fig:db} shows qualitative comparisons of reconstructed meshes on the Deep Blending dataset.
Our method recovers more accurate and complete surfaces, handling both dark and bright regions effectively.
Competing methods often exhibit noise, oversmoothing, or missing geometry, especially near object boundaries.
Results from GS-Pull are omitted due to its poor mesh quality and potential need for significant parameter tuning.
For other baselines, we use the same default parameters as used in their evaluations on the Mip-NeRF 360 dataset.

\begin{figure}[t]
    \centering
    \includegraphics[width=\linewidth]{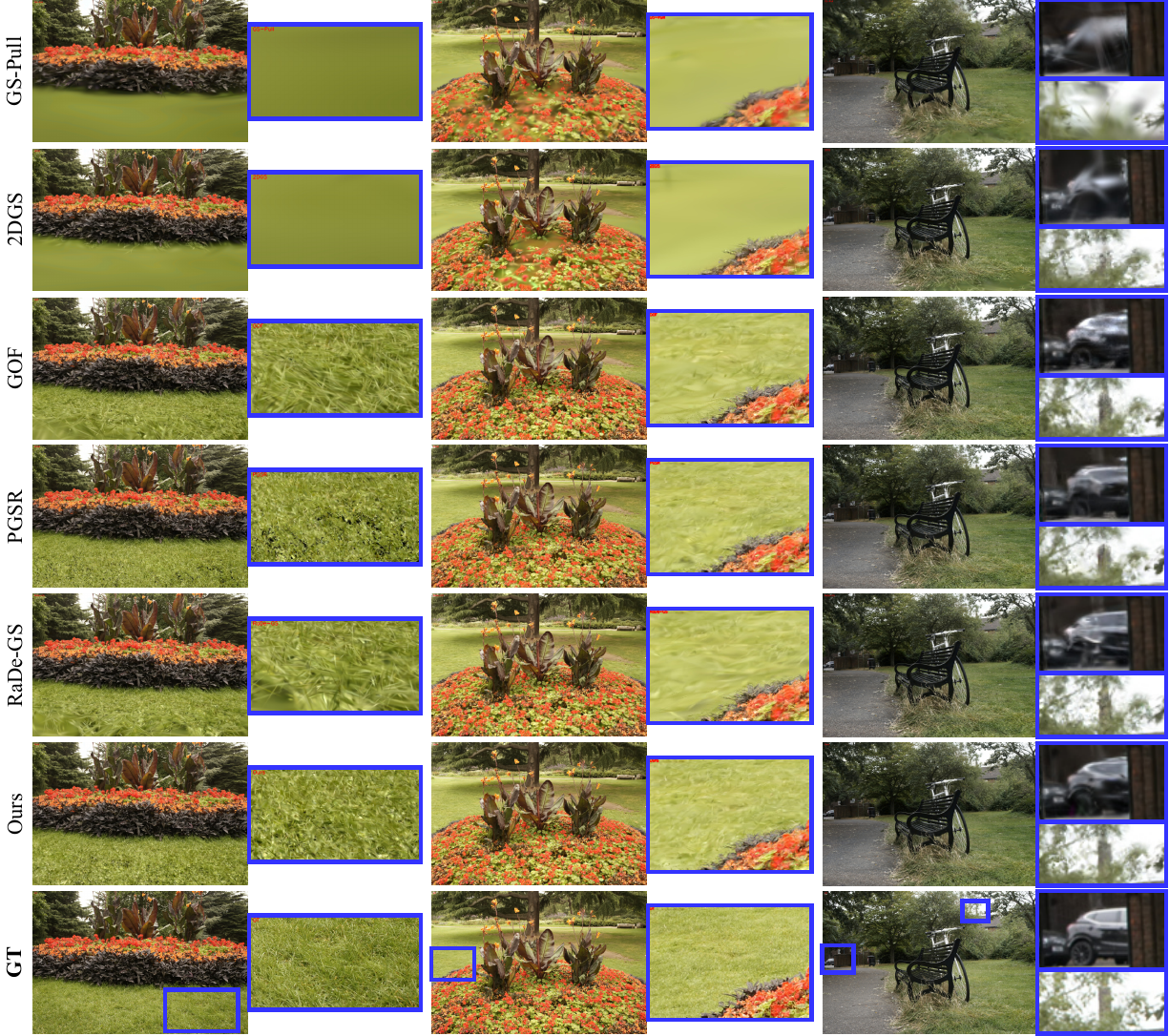}  \vspace{-0.6cm}
    \caption{
        Comparison of our method with prior Gaussian-based approaches on novel view synthesis on the Mip-NeRF 360 dataset.
        Our method produces high-quality renderings with clear details, while previous methods generate blurry results in complex regions such as grass, vegetation, and cars, as highlighted in the framed areas.
    }
    \label{fig:nvs}
    \vspace{-0.2cm}
\end{figure}

\begin{figure}[t]
    \centering
    \includegraphics[width=\linewidth]{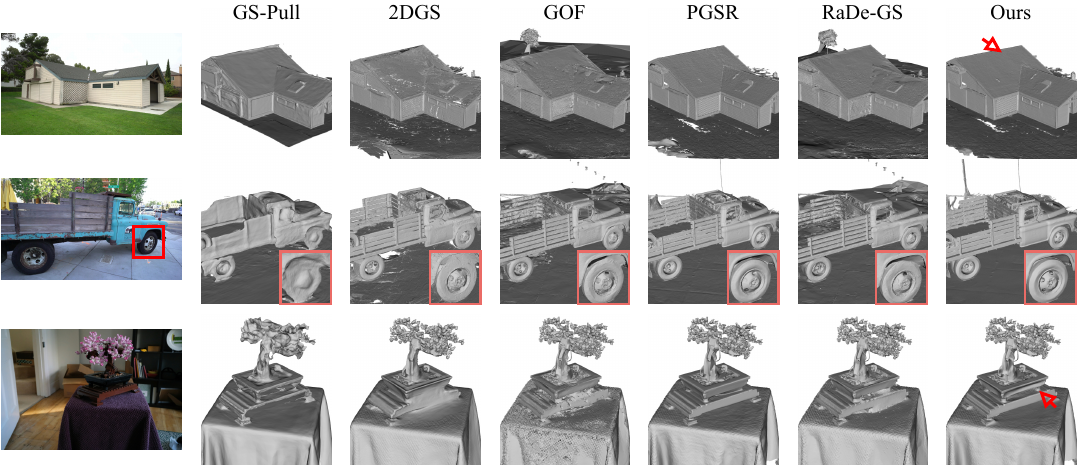}  \vspace{-0.6cm}
    \caption{
        Visual comparison of surface reconstruction results on the TNT (first two rows) and Mip-NeRF 360 (third row) datasets.
        Our method produces more complete and accurate surfaces with clearer object boundaries and fewer artifacts, effectively handling challenges such as complex lighting conditions and ambiguous geometric structures.
    }
    \label{fig:tnt_mip}
    % \vspace{-0.3cm}
\end{figure}

\begin{figure}[t]
    \centering
    \includegraphics[width=\linewidth]{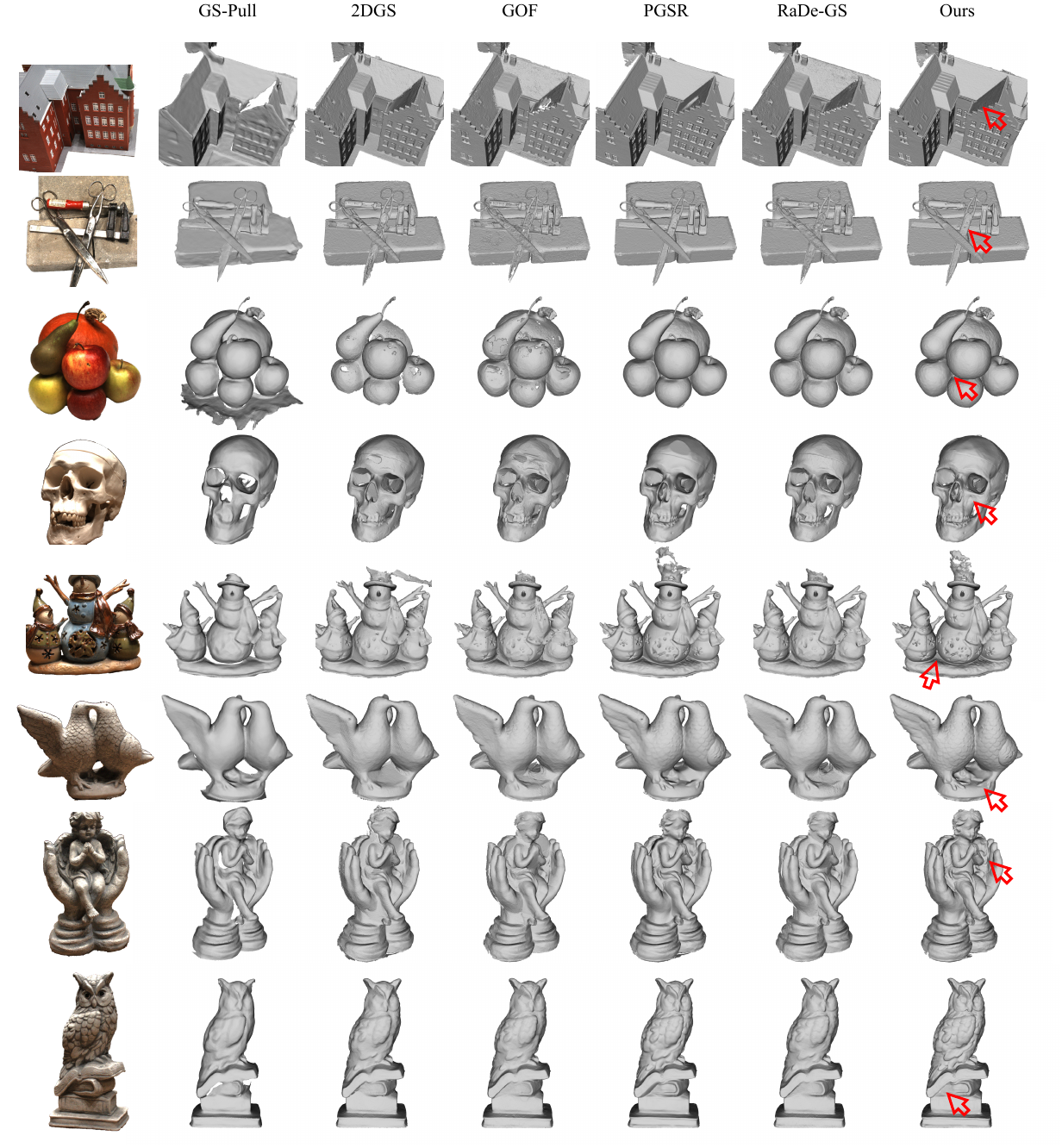}  \vspace{-0.6cm}
    \caption{
        Visual comparison of surface reconstruction results on the DTU dataset.
        Our method demonstrates superior capability in handling reflective materials and recovering fine-grained surface geometry.
        Please zoom in on the digital version for detailed inspection.
    }
    \label{fig:dtu}
    \vspace{-0.1cm}
\end{figure}

\subsection{More Experimental Results}

\textbf{Novel view synthesis.}
To provide a more comprehensive evaluation of novel view synthesis, we report per-scene quality metrics (PSNR, SSIM, and LPIPS) for Gaussian-based methods on the Mip-NeRF 360 dataset~\cite{barron2022mip}, as shown in Table~\ref{tab:mip360}.
All results are obtained from our own runs using the official implementations of prior methods, except for 3DGS, where we use the pretrained models provided by the authors to avoid training inconsistencies.
Our method consistently achieves the highest PSNR and SSIM scores on most scenes and has the best average scores.
Notably, it yields significant improvements in LPIPS across all scenes, highlighting its ability to capture high-frequency details and produce perceptually superior renderings.

Qualitative comparisons are shown in Fig.~\ref{fig:nvs}, where we visualize novel view synthesis results on the Mip-NeRF 360 dataset.
As highlighted in the boxed regions, our method produces sharper and more detailed renderings, particularly in complex areas such as grass and foliage, where competing methods often yield blurry outputs.

\textbf{Surface reconstruction.}
Fig.~\ref{fig:tnt_mip} presents additional qualitative comparisons of reconstructed surfaces on real-world indoor and outdoor scenes from the TNT~\cite{knapitsch2017tanks} and Mip-NeRF 360~\cite{barron2022mip} datasets.
Compared to baseline methods, our approach produces more complete and continuous meshes with well-preserved high-frequency details.
It also effectively avoids local minima, maintaining fine structures such as holes and sharp edges.
By contrast, 2DGS~\cite{huang20242d} and GOF~\cite{yu2024gaussian} often yield non-manifold meshes with broken topology, while GS-Pull~\cite{zhang2024neural}, which extracts surfaces from a learned SDF, tends to produce overly smooth geometry.

Additional comparisons of surface reconstruction on the DTU dataset~\cite{jensen2014large} are shown in Fig.~\ref{fig:dtu}.
Given the relatively simple geometry of the target objects, most methods achieve visually complete reconstructions.
However, the differences lie in the surface quality and detail.
Our method demonstrates superior performance in handling reflective surfaces and preserves fine-scale features, producing reconstructions that are both smoother and more accurate, thereby improving visual fidelity and geometric realism.

\begin{table}[t]
    \centering
    % \small
    \footnotesize
    \setlength{\tabcolsep}{1.5mm}
    \caption{
        Quantitative comparison on the Deep Blending dataset.
        The best results are highlighted as \colorbox{orange}{1st}, \colorbox{pink}{2nd} and \colorbox{yellow}{3rd}.
        $\ast$ indicates results copied from the original paper.
    }
    \vspace{-0.2cm}
    \label{tab:db}
    % \resizebox{\textwidth}{!}{
    \begin{tabular}{@{}lccccccccc}
        \toprule
        & \multicolumn{3}{c}{Dr Johnson}                            & \multicolumn{3}{c}{Playroom}                              & \multicolumn{3}{c}{\textbf{Mean}}                        \\
        & PSNR~$\uparrow$ & SSIM~$\uparrow$ & LPIPS~$\downarrow$    & PSNR~$\uparrow$ & SSIM~$\uparrow$ & LPIPS~$\downarrow$    & PSNR~$\uparrow$ & SSIM~$\uparrow$ & LPIPS~$\downarrow$   \\
        \cmidrule(r){1-1} \cmidrule(lr){2-4} \cmidrule(l){5-7} \cmidrule(l){8-10}

        3DGS~\cite{kerbl20233d}             & \sbest28.94 & \tbest0.896 & \sbest0.248 & 29.93       & 0.901       & 0.244       & \sbest29.43 & 0.898       & \sbest0.246 \\
        SuGaR$^\ast$~\cite{guedon2024sugar} & 28.71       & 0.889       & 0.273       & \tbest30.12 & 0.898       & 0.261       & \tbest29.41 & 0.893       & 0.267       \\
        2DGS~\cite{huang20242d}             & \tbest28.89 & \sbest0.898 & 0.259       & 29.88       & 0.901       & 0.259       & 29.38       & \tbest0.899 & 0.259       \\
        GS-Pull~\cite{zhang2024neural}      & 25.69       & 0.830       & 0.387       & 25.89       & 0.838       & 0.375       & 25.79       & 0.834       & 0.381       \\
        GOF~\cite{yu2024gaussian}           & 27.85       & 0.893       & 0.257       & \sbest30.16 & \sbest0.904 & \sbest0.242 & 29.01       & 0.899       & 0.250       \\
        RaDe-GS~\cite{zhang2024rade}        & 27.83       & \tbest0.896 & 0.257       & 30.04       & \best0.905  & \tbest0.243 & 28.94       & \sbest0.901 & 0.250       \\
        PGSR~\cite{chen2024pgsr}            & 28.61       & 0.891       & \tbest0.251 & 29.92       & \tbest0.903 & \tbest0.243 & 29.27       & 0.897       & \tbest0.247 \\
        Ours                                & \best29.10  & \best0.900  & \best0.241  & \best30.22  & \best0.905  & \best0.241  & \best29.66  & \best0.903  & \best0.241  \\

        \bottomrule
    \end{tabular} %}
    \vspace{-0.1cm}
\end{table}

\begin{table}[t]
    \centering
    \footnotesize
    \setlength{\tabcolsep}{1.6mm}
    \caption{
        Quantitative results on the Mip-NeRF 360 dataset.
        We report PSNR, SSIM, and LPIPS for each scene.
        All results are from our own runs using official code, except for 3DGS, where we use the authors' pre-trained models.
        The best results are highlighted as \colorbox{orange}{1st}, \colorbox{pink}{2nd} and \colorbox{yellow}{3rd}.
    }
    \vspace{-0.2cm}
    \label{tab:mip360}
    % \resizebox{\linewidth}{!}{
    \begin{tabular}{lccccccccc|c}
        \toprule
                                       & bicycle               & flowers     & garden      & stump       & treehill    & room        & counter     & kitchen     & bonsai      & \textbf{Mean} \\
        \midrule
        \textbf{PSNR}~$\uparrow$       & \multicolumn{10}{c}{}                                                                                                                                 \\
        3DGS~\cite{kerbl20233d}        & 25.17                 & 21.45       & 27.18       & 26.56       & 22.30       & \best31.34  & \best28.89  & 30.71       & \best31.98  & \tbest27.29   \\
        2DGS~\cite{huang20242d}        & \sbest24.71           & 21.03       & 26.61       & 26.11       & 22.27       & 30.72       & 28.08       & 30.30       & 31.24       & 26.79         \\
        GS-Pull~\cite{zhang2024neural} & 24.19                 & 20.56       & 26.08       & 25.24       & \best22.60  & \sbest30.84 & 26.41       & 26.06       & 29.10       & 25.68         \\
        RaDe-GS~\cite{zhang2024rade}   & 25.56                 & \sbest21.67 & \tbest27.39 & \sbest27.10 & 22.34       & \sbest30.84 & \tbest28.73 & \best31.26  & \tbest31.73 & \sbest27.40   \\
        GOF~\cite{yu2024gaussian}      & 25.44                 & \tbest21.59 & 27.27       & 26.93       & \sbest22.40 & 30.42       & 28.62       & 30.64       & 31.50       & 27.20         \\
        PGSR~\cite{chen2024pgsr}       & \tbest25.66           & 21.52       & \sbest27.49 & \tbest26.98 & 22.29       & 30.06       & 28.31       & \tbest30.80 & 31.55       & 27.18         \\
        Ours                           & \best25.89            & \best21.71  & \best27.69  & \best27.33  & \tbest22.38 & \tbest30.73 & \sbest28.77 & \sbest31.09 & \sbest31.93 & \best27.50    \\
        \midrule
        \textbf{SSIM}~$\uparrow$       & \multicolumn{10}{c}{}                                                                                                                                 \\
        3DGS~\cite{kerbl20233d}        & 0.762                 & 0.602       & 0.861       & 0.770       & \tbest0.631 & \tbest0.916 & \tbest0.905 & 0.923       & 0.938       & 0.812         \\
        2DGS~\cite{huang20242d}        & 0.729                 & 0.569       & 0.838       & 0.753       & 0.614       & 0.906       & 0.892       & 0.915       & 0.929       & 0.794         \\
        GS-Pull~\cite{zhang2024neural} & 0.660                 & 0.498       & 0.776       & 0.661       & 0.617       & 0.896       & 0.846       & 0.838       & 0.897       & 0.743         \\
        RaDe-GS~\cite{zhang2024rade}   & \sbest0.793           & \sbest0.640 & \tbest0.870 & \sbest0.801 & \sbest0.648 & \tbest0.916 & \tbest0.905 & \tbest0.924 & \tbest0.939 & \tbest0.826   \\
        GOF~\cite{yu2024gaussian}      & \tbest0.786           & 0.634       & 0.864       & 0.790       & 0.641       & 0.911       & 0.900       & 0.915       & 0.935       & 0.820         \\
        PGSR~\cite{chen2024pgsr}       & \sbest0.793           & \tbest0.635 & \sbest0.873 & \tbest0.798 & \best0.660  & \sbest0.927 & \sbest0.914 & \sbest0.932 & \sbest0.945 & \sbest0.831   \\
        Ours                           & \best0.803            & \best0.648  & \best0.877  & \best0.810  & \best0.660  & \best0.930  & \best0.919  & \best0.934  & \best0.948  & \best0.837    \\
        \midrule
        \textbf{LPIPS}~$\downarrow$    & \multicolumn{10}{c}{}                                                                                                                                 \\
        3DGS~\cite{kerbl20233d}        & 0.216                 & 0.341       & 0.115       & 0.219       & 0.328       & 0.223       & \tbest0.204 & \tbest0.130 & 0.208       & 0.220         \\
        2DGS~\cite{huang20242d}        & 0.276                 & 0.380       & 0.149       & 0.263       & 0.381       & 0.243       & 0.230       & 0.146       & 0.228       & 0.255         \\
        GS-Pull~\cite{zhang2024neural} & 0.331                 & 0.443       & 0.223       & 0.362       & 0.414       & 0.239       & 0.269       & 0.248       & 0.242       & 0.308         \\
        RaDe-GS~\cite{zhang2024rade}   & \sbest0.176           & 0.286       & \sbest0.103 & \sbest0.190 & \tbest0.279 & \tbest0.218 & 0.205       & \tbest0.130 & 0.204       & \tbest0.199   \\
        GOF~\cite{yu2024gaussian}      & \tbest0.182           & \tbest0.282 & \tbest0.109 & 0.197       & 0.282       & 0.221       & 0.205       & 0.137       & \tbest0.200 & 0.202         \\
        PGSR~\cite{chen2024pgsr}       & 0.186                 & \sbest0.264 & \sbest0.103 & \tbest0.192 & \sbest0.273 & \sbest0.180 & \sbest0.172 & \sbest0.113 & \sbest0.169 & \sbest0.184   \\
        Ours                           & \best0.169            & \best0.258  & \best0.098  & \best0.177  & \best0.254  & \best0.175  & \best0.164  & \best0.109  & \best0.163  & \best0.174    \\
        \bottomrule
    \end{tabular}  %}
    \vspace{-0.2cm}
\end{table}

\begin{table}[t]
    % \begin{wraptable}{r}{0.5\textwidth}
    \centering
    % \small
    \footnotesize
    \setlength{\tabcolsep}{1.5mm}
    \caption{
        Ablations on the TNT dataset.
    }
    \vspace{-0.2cm}
    \label{tab:ablation_supp}
    % \resizebox{0.92\columnwidth}{!}{
    \begin{tabular}{lccc}
        \toprule
                           & Precision~$\uparrow$ & Recall~$\uparrow$ & F1-score~$\uparrow$ \\
        \midrule
        $3\times3$         & 0.49                 & 0.58              & 0.52                \\
        $5\times5$         & 0.50                 & \textbf{0.60}     & \textbf{0.54}       \\
        $9\times9$         & \textbf{0.51}        & \textbf{0.60}     & \textbf{0.54}       \\
        Random cube init   & 0.44                 & 0.59              & 0.49                \\
        Random sphere init & 0.47                 & 0.59              & 0.51                \\
        $\tau \!=\! 0.001$ & 0.50                 & \textbf{0.60}     & 0.53                \\
        $\tau \!=\! 0.03$  & 0.50                 & 0.59              & 0.53                \\
        % vs=0.001           & 0.49                 & 0.60              & 0.53                \\
        % vs=0.002           & 0.51                 & 0.60              & 0.54                \\
        % vs=0.003           & 0.49                 & 0.58              & 0.52                \\
        \midrule
        Ours               & \textbf{0.51}        & \textbf{0.60}     & \textbf{0.54}       \\
        \bottomrule
    \end{tabular}  %}
    \vspace{-0.1cm}
\end{table}
% \end{wraptable}

%%%%%%%%%%%%%%%%%%%%%%%%%%%%%%%%%%%%%%%%%%%%%%%%%%%%%%%
\section{More Ablation Results}

To thoroughly evaluate the effectiveness and individual contributions of the components proposed in our method, we further conducted a series of ablation studies.
These experiments systematically replace or remove specific modules and vary key hyperparameters to assess their impact on overall performance.
The quantitative results are reported in Table~\ref{tab:ablation_supp}.

(1) \textit{Patch size in $\mathcal{L}_p$.}
In our multi-view photometric alignment loss $\mathcal{L}_p$, we use a default patch size of $7 \!\times\! 7$.
We also evaluate alternative patch sizes, including $3 \!\times\! 3$, $5 \!\times\! 5$, and $9 \!\times\! 9$.
As shown in Table~\ref{tab:ablation_supp}, smaller patches lead to suboptimal performance due to limited spatial context.
Larger patches, such as $9 \!\times\! 9$, offer no significant improvement while increasing computational cost.
This confirms that our chosen patch size achieves a good balance between accuracy and efficiency.

(2) \textit{Gaussian Initialization.}
Similar to prior works, we use COLMAP~\cite{schoenberger2016sfm} to generate a sparse point cloud from the input images, which serves as the initialization for our 3D Gaussians.
To evaluate the impact of initialization quality, we experiment with randomly generated point clouds using two alternative strategies: (i) uniform sampling within a cube, and (ii) sampling points on a spherical surface. % surrounding the scene.
As shown in Table~\ref{tab:ablation_supp}, both alternatives degrade the final reconstruction quality compared to COLMAP initialization.
However, the spherical initialization performs better than the cube sampling, likely due to its more uniform coverage and approximate enclosure of the scene geometry.

(3) \textit{Threshold in $\mathcal{L}_{ns}$.}
Our normal-based geometric alignment module incorporates a smoothing term $\mathcal{L}_{ns}$, with a threshold set to $0.01$ by default.
This threshold helps preserve sharp edges while reducing noise in low-curvature regions.
We conduct ablations by testing alternative values, including $0.001$ and $0.03$.
The results show that excessively small values (\eg, $0.001$) overly constrain normal variation, leading to surface oversmoothing, while large values (\eg, $0.03$) reduce the loss's regularization effect.
Our default setting ($0.01$) strikes a good balance and achieves the best performance.

\textbf{The run-time cost of each loss.}
We provide the runtime of each loss ablation on the TNT dataset in Table~\ref{tab:ablation_time}.
With all losses enabled, the training takes $20.9$ minutes on RTX 4090.
Removing $\mathcal{L}_f$ (feature alignment) reduces training time to $15.9$ minutes, yielding a $\sim5$-minute saving, while still preserving strong performance.
By comparison, removing $\mathcal{L}_p$ (photometric alignment) saves more time ($8.3$ min) but sacrifices more reconstruction quality.
When only $\mathcal{L}_I$ (image reconstruction) is used, our method has a lower time cost than the original 3DGS ($5.9$ min \vs $7.5$ min) and provides better F1-score results than 3DGS ($0.13$ \vs $0.09$).

\begin{table}[t]
    \centering
    \footnotesize
    \caption{
        Ablation study on different loss terms.
    }
    \vspace{-0.2cm}
    \label{tab:ablation_time}
    \begin{tabular}{l|c|cccccc}
        \toprule
                                & 3DGS & Only $\mathcal{L}_I$ & w/o $\mathcal{L}_{nc}$ & w/o $\mathcal{L}_{ns}$ & w/o $\mathcal{L}_{p}$ & w/o $\mathcal{L}_{f}$ & Full          \\
        \midrule
        F1-score $\uparrow$     & 0.09 & 0.13                 & 0.52                   & 0.51                   & 0.50                  & 0.53                  & \textbf{0.54} \\
        Time (min) $\downarrow$ & 7.5  & \textbf{5.9}         & 20.3                   & 17.1                   & 12.6                  & 15.9                  & 20.9          \\
        Time Gap                & -    & 15.0                 & 0.6                    & 3.8                    & 8.3                   & 5.0                   & 0.0           \\
        \bottomrule
    \end{tabular}
    \vspace{-0.1cm}
\end{table}

%%%%%%%%%%%%%%%%%%%%%%%%%%%%%%%%%%%%%%%%%%%%%%%%%%%%%%%
\section{More Discussion}

\textbf{More clarification of loss $\mathcal{L}_f$.}
We agree that $\mathcal{L}_f$ brings modest improvements on the TNT dataset.
$\mathcal{L}_f$ improves general robustness and is beneficial in scenes with consistent lighting or texture.
It improves robustness under low-texture or lighting-variant scenes, which is not fully reflected by the F1-score metric alone.
Removing it often leads to over-smoothed meshes in challenging regions.
Moreover, the contribution of the feature alignment loss $\mathcal{L}_f$ is more significant on the DTU dataset: removing $\mathcal{L}_f$ increases Chamfer distance from $0.49$ to $0.52$.
We attribute this to dataset characteristics.
DTU scenes benefit more from learned feature-level consistency due to cleaner lighting and smoother structure.
TNT contains diverse indoor/outdoor scenes and severe lighting variation.
The expressive power of off-the-shelf image features is limited, which may partially underutilize $\mathcal{L}_f$.
We plan to explore stronger feature extractors in future work.

\textbf{Clarification of performance gains and training time.}
(a) While our training time is longer than that of vanilla 3DGS, the added cost stems from the newly introduced geometry supervision and view-alignment mechanisms, which directly improve surface accuracy.
These components are essential for multi-view reconstruction, especially in challenging scenes.
While the gains over the latest baselines may appear modest in some scenes, these differences represent noticeable geometric improvements.
(b) Moreover, increased time is a common trade-off among geometry-enhanced 3DGS variants, where introducing geometric priors and view constraints improves reconstruction quality at the cost of runtime (see Tables 1 and 2 of main paper).
The original 3DGS remains the fastest due to its lightweight constraints, but also exhibits the least accurate geometry.
In future work, we plan to explore adaptive Gaussian pruning to further reduce training cost without sacrificing accuracy.
(c) Our goal of this work is not to accelerate or compress 3DGS, but to enhance its geometric representation.
On DTU, we reduce Chamfer distance from $1.96$ to $0.49$; on TNT, we raise F1-score from $0.09$ to $0.54$, a substantial improvement over 3DGS; on Mip-NeRF 360, we also observe consistent gains across all metrics.
As shown in Table~\ref{tab:ablation_time}, when only using edge-aware image reconstruction loss $\mathcal{L}_I$ on TNT, our method runs faster than 3DGS ($5.9$ min \vs $7.5$ min) and achieves a higher F1-score ($0.13$ \vs $0.09$).
(d) In terms of practicality, our training time on TNT ($\sim20$ min) is production-viable for many offline applications.
In contrast, neural implicit SDF methods often require $10+$ hours of training.
We believe that accuracy is often prioritized over marginal runtime gains in such scenarios, with the rapid advancement of hardware.

\textbf{More explanation of the ablations.}
In all ablation studies, we adopt a consistent and controlled protocol to ensure fair comparisons:
(a) When ablating a loss term (\eg, $\mathcal{L} _ {p}$ or $\mathcal{L} _ {f}$), we drop the corresponding term entirely from the optimization objective.
(b) The remaining loss weights are kept unchanged, and we do not re-normalize or re-scale other terms to preserve consistent training dynamics.
(c) All experiments are conducted with the same number of training iterations and identical optimization settings (\eg, learning rate, batch size, data split).
(d) For the hyperparameter ablations, we similarly vary only the parameter under study (\eg, source view count, threshold value), while keeping all other parameters and training settings fixed.
This ensures that any performance change can be attributed directly to the presence or absence of the specific loss being tested.

\textbf{The difference from PGSR.}
Firstly, we would like to clarify that our method is not a direct extension of PGSR with feature alignment added on top.
Instead, our framework introduces a new combination of single-view alignment (including edge-aware image reconstruction and normal-based geometry supervision) and multi-view alignment (photometric and feature alignment), which are structurally and conceptually different from PGSR.
As shown in the ablations of Table 4 of main paper, even without feature alignment, our method already achieves better performance than PGSR.
This demonstrates that our core design, particularly single-view and multi-view photometric supervision, is effective.
The additional feature alignment further improves the results, validating our design choices.
Even when the improvement appears numerically small on the TNT dataset, reaching the same or slightly higher performance than PGSR still represents a state-of-the-art level.
Moreover, on the DTU dataset, the contribution of the feature alignment loss $\mathcal{L}_f$ is more significant.
Removing $\mathcal{L}_f$ increases Chamfer distance from $0.49$ to $0.52$, which is still better than PGSR’s reported $0.53$.
This shows that our method offers consistent geometric improvement across datasets.
Finally, our method also outperforms PGSR on the Mip-NeRF 360 dataset for novel view synthesis, as shown in Table 3 of main paper.
We achieve better scores on all metrics, indicating that our method is not merely comparable to PGSR but surpasses it in reconstruction/synthesis quality and generalizability across varied benchmarks.

Our Multi-View Photometric Alignment differs from PGSR’s in several key aspects:
(a) Optimization complexity: PGSR couples its photometric consistency loss with geometric consistency regularization, which minimizes forward-backward reprojection error of neighboring views, resulting in more variables participating in network backpropagation and a more complex gradient computation.
In contrast, our method simplifies the backpropagation path by aligning Gaussian orientations first and then applying image/feature-level constraints, which leads to better optimization efficiency.
(b) Multi-view formulation: PGSR computes photometric consistency between one reference view and one source view.
Our framework uses one reference view and multiple source views (three by default, varied in ablation), which introduces richer multi-view constraints and improves geometric supervision.
(c) Gaussian flattening: PGSR flattens 3D Gaussians using scale regularization of the Gaussian ellipsoid to mimic planar surfaces.
In our method, we found this strategy ineffective through ablation in Table 4 of main paper, and therefore designed our supervision differently.
(d) Efficiency and performance: As shown in Table 2 of main paper, our method achieves better reconstruction quality across more scenes on real outdoor scenes of the TNT dataset while requiring less training time than PGSR.
(e) Occlusion modeling: We explicitly define a visibility term and occlusion weight to mitigate the effect of outlier pixels caused by occlusion or misalignment.
We give their motivation and clearly show the derivation of their formulas in the method section.

%%%%%%%%%%%%%%%%%%%%%%%%%%%%%%%%%%%%%%%%%%%%%%%%%%%%%%%
\section{Limitations and Future Works}

Despite achieving strong performance in both surface reconstruction and novel view synthesis, our method has several limitations that suggest promising directions for future work.
First, our model assumes known camera poses as input, which may not always be available in real-world scenarios, especially when the number of input views is extremely sparse.
Removing this requirement by exploring pose-free approaches~\cite{ye2024no,zhang2025flare} is a compelling future direction that would increase the applicability of our method in unconstrained settings.
Moreover, our method assumes a moderate number of input views.
For extremely sparse-view settings, the multi-view consistency constraints may become less effective, reducing reconstruction quality.
Second, the number of Gaussians grows significantly with the number of input views, potentially limiting the scalability of our method in large-scale or densely captured scenes.
Designing more compact or adaptive Gaussian representations, or incorporating efficient pruning and regularization strategies, could help improve scalability without compromising reconstruction quality.
Third, our method currently relies on a pre-trained feature extraction network to guide geometric alignment and ensure robustness under varying lighting and viewpoints.
This dependence introduces limitations, particularly when the pre-trained features are suboptimal for specific domains or scenes.
Exploring more reliable, possibly self-supervised, feature learning strategies or reducing reliance on external networks altogether would further improve robustness and generalization.
Addressing these limitations can pave the way for more flexible, efficient, and generalizable Gaussian-based surface reconstruction frameworks.

%%%%%%%%%%%%%%%%%%%%%%%%%%%%%%%%%%%%%%%%%%%%%%%%%%%%%%%
\section{Potential Negative Social Impacts}

While our method advances the state-of-the-art in surface reconstruction and novel view synthesis using 3D Gaussian Splatting, it is important to consider its potential negative social impacts.
First, high-fidelity 3D reconstruction from multi-view imagery could be misused to reconstruct environments or individuals without consent.
For instance, when applied to personal photos or public surveillance footage, our method may enable unauthorized digital replication of private spaces or identities.
We strongly advocate for the ethical and consensual collection and use of input data.
Second, enhanced geometry reconstruction and photorealistic novel view synthesis could be integrated into pipelines for generating synthetic scenes or manipulating real-world data. This could potentially contribute to the creation of deceptive media, raising concerns about misinformation, impersonation, or forgery. Preventative measures such as watermarking and provenance tracking should be considered in downstream applications.
Third, although more efficient than dense volumetric methods, our approach still requires significant GPU resources for training and inference.
As the field moves toward real-time and scalable 3D vision, energy consumption and environmental sustainability must remain part of the design considerations.

% %%%%%%%%%%%%%%%%%%%%%%%%%%%%%%%%%%%%%%%%%%%%%%%%%%%%%%%%%%%%
% %%% Bibliography
% {
% \small
% \bibliographystyle{ieeenat_fullname}
% \bibliography{egbib}
% }

% \end{document}

\end{document}